\documentclass[journal]{IEEEtran}
\usepackage{cite}
\usepackage{graphicx}
\usepackage{amsmath}
\usepackage{array}
\usepackage{amssymb}
\usepackage{booktabs}
\usepackage{bbding}
\usepackage{subfigure}
\usepackage{textcomp}
\usepackage{fixltx2e}
\usepackage{url}
\usepackage{multirow}
\usepackage{epsfig}
\usepackage{threeparttable}
\def\ie{{\em i.e.}}
\def\eg{{\em e.g.}}
\def\etal{{\em et al.}}
\begin{document}
\title{CASIA-SURF: A Large-scale Multi-modal Benchmark for Face Anti-spoofing}
\author{Shifeng~Zhang$^*$,
        Ajian~Liu$^*$,
        Jun~Wan$^\dag$,~\IEEEmembership{Senior Member,~IEEE,}
        Yanyan~Liang,~\IEEEmembership{Senior Member,~IEEE,}
        Guogong~Guo,~\IEEEmembership{Senior Member,~IEEE,}
        Sergio~Escalera,
        Hugo~Jair~Escalante,
        Stan~Z.~Li,~\IEEEmembership{Fellow,~IEEE}
\thanks{Shifeng Zhang and Jun Wan are with the National Laboratory of Pattern Recognition (NLPR), Institute of Automation Chinese Academy of Sciences (CASIA) and University of Chinese Academy of Sciences (UCAS), Beijing, China (e-mail: \{shifeng.zhang, jun.wan\}@nlpr.ia.ac.cn).}
\thanks{Ajian Liu and Yanyan Liang are with the Macau University of Science and Technology (MUST), Macau, China (e-mail: ajianliu92@gmail.com, yyliang@must.edu.mo).}
\thanks{Guodong Guo is with the Institute of Deep Learning, Baidu Research and National Engineering Laboratory for Deep Learning Technology and Application (e-mail: guoguodong01@baidu.com).}
\thanks{Sergio Escalera is with the Universitat de Barcelona (UB) and Computer Vision Center (CVC), Barcelona, Catalonia, Spain (e-mail: sergio@maia.ub.es).}
\thanks{Hugo Jair Escalante is with  Instituto Nacional de Astrof\'isica, \'Optica y Electr\'onica, Puebla,  Mexico and  Computer Science Department at  CINVESTAV-Zacatenco, Mexico (e-mail: hugojair@inaoep.mx).}
\thanks{Stan Z. Li is with the Westlake University and Institute of Automation Chinese Academy of Sciences (CASIA) (e-mail: Stan.ZQ.Li@westlake.edu.cn).}
\thanks{$^*$Equal contribution. \quad $^\dag$ Corresponding author.}
}

\markboth{Journal of IEEE Transactions on Biometrics, Behavior, and Identity Science,~Vol.~xx, No.~x, xxx~2020}%
{Zhang \MakeLowercase{~\etal}: CASIA-SURF: A Large-scale Multi-modal Benchmark for Face Anti-spoofing}

\maketitle

\begin{abstract}
Face anti-spoofing is essential to prevent face recognition systems from a security breach. Much of the progresses have been made by the availability of face anti-spoofing benchmark datasets in recent years. However, existing face anti-spoofing benchmarks have limited number of subjects ($\le\negmedspace170$) and modalities ($\leq\negmedspace2$), which hinder the further development of the academic community. To facilitate face anti-spoofing research, we introduce a large-scale multi-modal dataset, namely CASIA-SURF, which is the largest publicly available dataset for face anti-spoofing in terms of both subjects and modalities. Specifically, it consists of $1,000$ subjects with $21,000$ videos and each sample has $3$ modalities (\ie, RGB, Depth and IR). We also provide comprehensive evaluation metrics, diverse evaluation protocols, training/validation/testing subsets and a measurement tool, developing a new benchmark for face anti-spoofing. Moreover, we present a novel multi-modal multi-scale fusion method as a strong baseline, which performs feature re-weighting to select the more informative channel features while suppressing the less useful ones for each modality across different scales. Extensive experiments have been conducted on the proposed dataset to verify its significance and generalization capability. The dataset is available at \url{https://sites.google.com/qq.com/face-anti-spoofing/welcome/challengecvpr2019?authuser=0}.
\end{abstract}


\IEEEpeerreviewmaketitle

\section{Introduction}
\IEEEPARstart{F}{ace} anti-spoofing aims to determine whether the captured face from a face recognition system is real or fake. With the development of deep Convolutional Neural Networks (CNNs), face recognition~\cite{zhang2019refineface, wang2019mis, zhang2019single, chi2018selective, zhang2019faceboxes, wang2018ensemble, zhang2019improved, zhang2018detecting} has achieved near-perfect recognition performance and already has been applied in our daily life, such as phone unlock, access control and face payment. However, these face recognition systems are prone to be attacked in various ways including print attack, video replay attack and 2D/3D mask attack, causing the recognition result to become unreliable. Therefore, face Presentation Attack Detection (PAD)~\cite{Boulkenafet2016Face,Boulkenafet2017Face} is a vital step to ensure that face recognition systems are in a safe reliable condition.

\begin{figure}[t]
\centering
\includegraphics[width=0.915\linewidth]{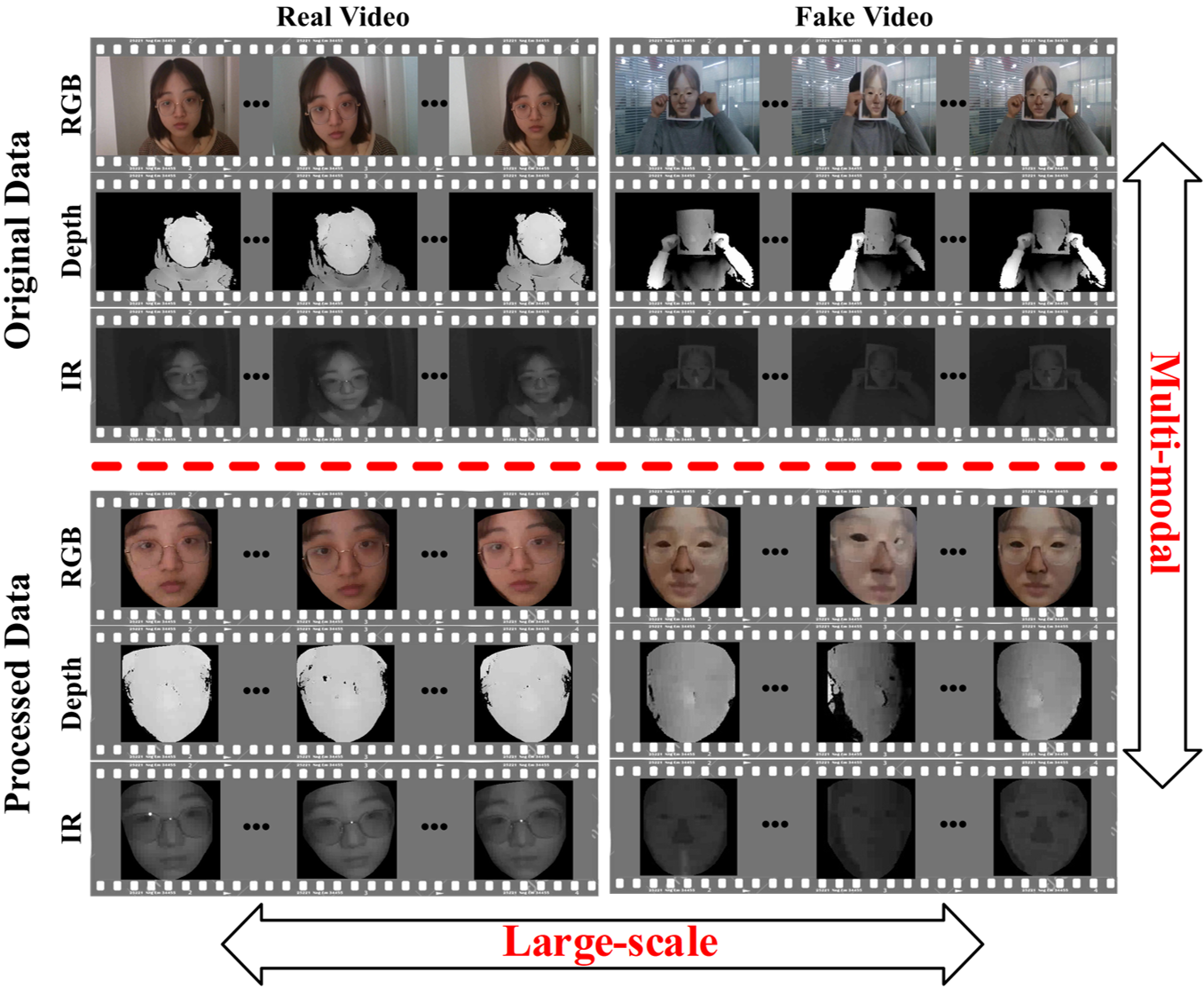}
\caption{The CASIA-SURF dataset. It is a large-scale and multi-modal dataset for face anti-spoofing, consisting of $492,522$ images with $3$ modalities (\ie, RGB, Depth and IR).}
\label{fig:dataset}
\end{figure}

\begin{table*}
\renewcommand\arraystretch{1.15}
\centering
\setlength{\tabcolsep}{9pt}
\caption{Comparison of the public face anti-spoofing datasets ($*$ indicates this dataset only contains images, not video clips, $\star$ is short for Seek Thermal Compact PRO sensor, $-$ indicates that this team is not counted).}
\footnotesize{
\begin{tabular}{|c|c|c|c|c|c|c|c|}
\hline
Dataset &Year &\# subjects & \# videos & Camera & Modal types&  Spoof attacks \\
\hline
\hline
Replay-Attack~\cite{Chingovska_BIOSIG-2012} & 2012 & 50&   1,200  & VIS& RGB&  Print, 2 Replay\\
\hline
CASIA-MFSD~\cite{Zhang2012A}                & 2012 & 50&   600   & VIS& RGB&  Print, Replay\\
\hline
3DMAD~\cite{ERDOGMUS_BTAS-2013}             & 2013 & 17&   255    & VIS/Kinect&   RGB/Depth&     3D Mask\\
\hline
I$^{2}$BVSD~\cite{Dhamecha2014Recognizing}  & 2013 & 75&   681$^{*}$ &VIS/Thermal &RGB/Heat&     3D Mask\\
\hline
GUC-LiFFAD~\cite{RaghavendraPresentation} & 2015 & 80&   4,826   &LFC &LFI& 2 Print, Replay \\
\hline
MSU-MFSD~\cite{Wen2015Face}                 & 2015 & 35&   440    & Phone/Laptop& RGB&  Print, 2 Replay\\
\hline
Replay-Mobile~\cite{Costa2016The}           & 2016 & 40&   1,030   & VIS&  RGB&  Print, Replay\\
\hline
3D Mask~\cite{liu20163d}                    & 2016 & 12& 1,008     & VIS&  RGB&  3D Mask \\
\hline
Msspoof~\cite{msspoof-2015}                 & 2016 & 21&   4,704$^{*}$ & VIS/NIR &RGB/IR  & Print\\
\hline
SWIR~\cite{steiner2016reliable} & 2016 & 5 & 141$^{*}$ & VIS/M-SWIR & RGB/4 SWIR bands & Print, 3D Mask \\
\hline
BRSU~\cite{HolgerDesign} & 2016 & 50+ &$-$ & VIS/AM-SWIR & RGB/4 SWIR bands & Print, 3D Mask \\
\hline
EMSPAD~\cite{raghavendra2017vulnerability} & 2017 & 50   & 14,000$^{*}$
& SpectraCam$^{TM}$    & 7 bands    & 2 Print \\
\hline
SMAD~\cite{manjani2017detecting} & 2017 & $-$   & 130 & VIS    & RGB    & 3D Mask \\
\hline
MLFP~\cite{agarwal2017face}  & 2017    & 10      & 1,350
& VIS/NIR/Thermal   &RGB/IR/Heat    &2D/3D Mask \\
\hline
Oulu-NPU~\cite{Boulkenafet2017OULU}         & 2017 & 55&   5,940 & VIS&          RGB&  2 Print, 2 Replay\\
\hline
SiW~\cite{Liu2018Learning}                  & 2018 & 165&  4,620 & VIS&          RGB&  2 Print, 4 Replay\\
\hline
WMCA~\cite{george2019biometric} & 2019 & 72 & 6,716
& RealSense/STC-PRO$^{\star}$ & RGB/Depth/IR/Thermal & 2 Print, Replay, 2D/3D Mask \\
\hline
\hline
\textbf{CASIA-SURF} & 2018 & \textbf{1,000} & \textbf{21,000} &  RealSense & RGB/Depth/IR &  Print, Cut\\
\hline
\end{tabular}
}
\label{table:datasets}
\end{table*}

In recent years, face PAD algorithms~\cite{Jourabloo2018Face,Liu2018Learning} have achieved great performances. One of the key points of this success is the availability of face anti-spoofing benchmark datasets~\cite{Boulkenafet2017OULU,Chingovska_BIOSIG-2012,Costa2016The,Liu2018Learning,Wen2015Face,Zhang2012A}. However, there are several shortcomings in the existing datasets as follows:
\begin{itemize}
\setlength{\itemsep}{1.0pt}
\item \textbf{Number of subjects is limited.} Compared to the large existing image classification \cite{deng2009imagenet} and face recognition \cite{yi2014learning} datasets, face anti-spoofing datasets have less than $170$ subjects and $60,00$ video clips as shown in Table~\ref{table:datasets}. The limited number of subjects is not representative of the requirements of real applications.
\item \textbf{Number of modalities is limited.} As shown in Table~\ref{table:datasets}, most of the existing datasets only consider a single modality (\eg, RGB). For these existing available multi-modal datasets~\cite{ERDOGMUS_BTAS-2013,msspoof-2015}, they are very scarce including no more than 21 subjects.
\item \textbf{Evaluation metrics are not comprehensive enough.} How to compute the performance of algorithms is an open issue in face anti-spoofing. Many works~\cite{Liu2018Learning,Jourabloo2018Face,Boulkenafet2017OULU,Costa2016The} adopt the Attack Presentation Classification Error Rate (APCER), the Normal Presentation Classification Error Rate (NPCER) and the Average Classification Error Rate (ACER) as the evaluation metric, in which APCER and NPCER are used to measure the error rate of fake or live samples, and ACER is the average of APCER and NPCER scores. However, in real applications, one may be more concerned about the false positive rate, \ie, attacker is treated as real/live one. These aforementioned metrics can not meet this need. 
\item \textbf{Evaluation protocols are not diverse enough.} All the existing face anti-spoofing datasets only provide within-modal evaluation protocols. To be more specific, algorithms trained in a certain modality can only be evaluated in the same modality, which limits the diversity of face anti-spoofing research.
\end{itemize}

To deal with these aforementioned drawbacks, we introduce a large-scale multi-modal face anti-spoofing dataset, namely CASIA-SURF, which consists of $1,000$ subjects and $21,000$ video clips with $3$ modalities (RGB, Depth, IR). It has $6$ types of photo attacks combined by multiple operations, \eg, cropping, bending the print paper and stand-off distance. Some samples and other detailed information of our dataset are shown in Fig.~\ref{fig:dataset} and Table~\ref{table:datasets}. Comparing to these existing face anti-spoofing datasets, the proposed dataset has four main advantages as follows:
\begin{itemize}
\setlength{\itemsep}{1.0pt}
\item \textbf{The most subjects.} The proposed dataset is the largest one in term of number of subjects, which is more than $6\times$ boosted compared with previous challenging face anti-spoofing dataset like Spoof in the Wild (SiW)~\cite{Liu2018Learning}.
\item \textbf{The most modalities.} Our CASIA-SURF is the only dataset that provides three modalities (\ie, RGB, Depth and IR), and the other datasets have up to two modalities.
\item \textbf{The most comprehensive evaluation metrics.} Inspired by face recognition~\cite{liu2017sphereface,wang2018support}, we introduce the Receiver Operating Characteristic (ROC) curve for our large-scale face anti-spoofing dataset in addition to the commonly used evaluation metrics. The ROC curve can be used to select a suitable trade off threshold between the False Positive Rate (FPR) and the True Positive Rate (TPR) according to the requirements of a given real application.
\item \textbf{The most diverse evaluation protocols.} In addition to the within-modal evaluation protocols, we also provide the cross-modal evaluation protocols in our dataset, in which algorithms trained in one modality will be evaluated in other modalities. It allows the academic community to explore new issues.
\end{itemize}

Besides, we present a novel multi-modal multi-scale fusion method as a strong baseline to conduct extensive experiments on the proposed dataset. Our new fusion method performs feature re-weighting to select the more informative channel features while suppressing the less useful ones for each modality across different scales. To sum up, the contributions of this paper are three-fold:
\begin{itemize}
\setlength{\itemsep}{1.0pt}
\item Presenting a large-scale multi-modal face anti-spoofing dataset with $1,000$ subjects and $3$ modalities.
\item Introducing a new multi-modal multi-scale fusion method to effectively merge the involved three modalities across different scales.
\item Conducting extensive experiments on the proposed CASIA-SURF dataset to verify its significance and generalization capability.
\end{itemize}

Preliminary results of this work have been published in \cite{DBLP:conf/cvpr/abs-1812-00408}. The current work has been improved and extended from the conference version in several important aspects. (1) We provide the cross-modal evaluation protocols in our dataset for the academic community to explore new issues. (2) We improve the multi-modal fusion method in our previous work from one scale to multiple scales for better performance. (3) Some additional experiments are conducted and we noticeably improve the accuracy of the baseline in our previous work. (4) All sections are rewritten with more details, more references and more analysis to have a more elaborate presentation.

\section{Related work}
Face anti-spoofing has made great progress with the proposal of new datasets in recent years. This section first summarizes the existing face anti-spoofing datasets and then reviews some representative methods

\subsection{Dataset}
Most of existing face anti-spoofing datasets only contain the RGB modality, including Replay-Attack~\cite{Chingovska_BIOSIG-2012}, CASIA-FASD~\cite{Zhang2012A} and SiW~\cite{Liu2018Learning}. With the popularity of face recognition in mobile phones, there are also some RGB datasets recorded by replaying face video with smartphone, such as MSU-MFSD~\cite{Wen2015Face}, Replay-Mobile~\cite{Costa2016The} and OULU-NPU~\cite{Boulkenafet2017OULU}.

As attack techniques are constantly upgraded, some new types of attacks have emerged, \eg, 3D~\cite{ERDOGMUS_BTAS-2013} and silicone masks~\cite{Bhattacharjee_BTAS2018_2018}. These attacks are more realistic than traditional 2D attacks and the drawbacks of visible cameras are revealed. Fortunately, some new sensors have been introduced to provide more possibilities for face PAD methods, such as depth, muti-spectral and infrared light cameras. Kim~\etal~\cite{Kim2009Masked} introduce a new dataset to distinguish between facial skin and mask materials by exploiting their reflectance. Kose~\etal~\cite{Kose2013Countermeasure} propose a 2D+3D face mask attack dataset to study the effects of mask attacks. 3DMAD~\cite{ERDOGMUS_BTAS-2013} is recorded using Microsoft Kinect sensor and consists of Depth and RGB modalities with 3D masks. Another multi-modal dataset is Msspoof~\cite{msspoof-2015}, containing visible and near-infrared images of real accesses and printed spoofing attacks with $\leq21$ objects.

However, existing face PAD datasets have two main limitations: 1) They have limited number of subjects and samples, resulting in a potential over-fitting risk; 2) Most of existing datasets only include the RGB modality, causing substantial failures when facing new types of attacks (\eg, 3D mask). 

\subsection{Method}
Previous face PAD works~\cite{Pan2007Eyeblink,wang2009face,kollreider2008verifying,Bharadwaj2013Computationally} attempt to detect the evidence of liveness (\eg, eye-blinking). Some works are based on contextual~\cite{Pan2011Monocular,Komulainen2014Context} and moving~\cite{Wang2013Face,De2012Moving,Kim2013Face} information. To improve the robustness to illumination variation, some algorithms adopt HSV and YCbCr color spaces~\cite{Boulkenafet2016Face,Boulkenafet2017Face}, as well as Fourier spectrum~\cite{Li2004Live}. All of these methods use handcrafted features, such as LBP~\cite{chingovska2012effectiveness,Yang2013Face,Maatta2012Face}, HoG~\cite{Yang2013Face,Maatta2012Face,schwartz2011face} and GLCM~\cite{schwartz2011face}. They achieve a relatively satisfactory performance on small public face anti-spoofing datasets.

Some fusion methods have been proposed to obtain a more general countermeasure effective against a variation of attack types. Tronci \etal~\cite{tronci2011fusion} propose a linear fusion of frame and video analysis. Schwartz \etal~\cite{schwartz2011face} introduce feature level fusion based on a set of low-level feature descriptors. Other works~\cite{Pereira2013Can,komulainen2013complementary} obtain an effective fusion scheme by measuring the independence level of two anti-counterfeiting systems. However, they only focus on score or feature level, not modality level, due to the lack of multi-modal datasets.

CNN-based methods~\cite{feng2016integration,li2016original,Patel2016Secure,yang2014learn, Liu2018Learning,Jourabloo2018Face} have been presented recently. They treat face PAD as a binary classification and achieve remarkable improvements. Liu~\etal~\cite{Liu2018Learning} design a network to leverage Depth map and rPPG signal as supervision. Amin~\etal~\cite{Jourabloo2018Face} solve the face anti-spoofing by inversely decomposing a spoof face into the live face and the spoof noise pattern. However, they exhibit a poor generalization ability due to the over-fitting to training data, even adopting transfer learning to train a CNN model~\cite{li2016original,Patel2016Secure} from ImageNet~\cite{deng2009imagenet}. These works show the need of a larger PAD dataset.

\section{CASIA-SURF dataset}
Existing datasets involve a limited number of subjects and modalities, which severely impedes the development of face PAD with higher recognition to be applied in problems, such as face payment or unlock. In order to address these aforementioned limitations, we collect a new large-scale and multi-modal face PAD dataset namely CASIA-SURF. To the best our knowledge, the proposed dataset is currently the largest face anti-spoofing dataset, containing $1,000$ Chinese people in $21,000$ videos with three modalities (RGB, Depth, IR). Another motivation for creating this dataset, beyond pushing the further research of face anti-spoofing, is to explore the performance of recent face anti-spoofing methods when considering a large amount of data. In this section, we will give the detailed introduction of the proposed dataset, including acquisition detail, attack type, data preprocessing, statistics description, evaluation metric and protocol.

\begin{figure}[b]
\centering
\includegraphics[width=1\linewidth]{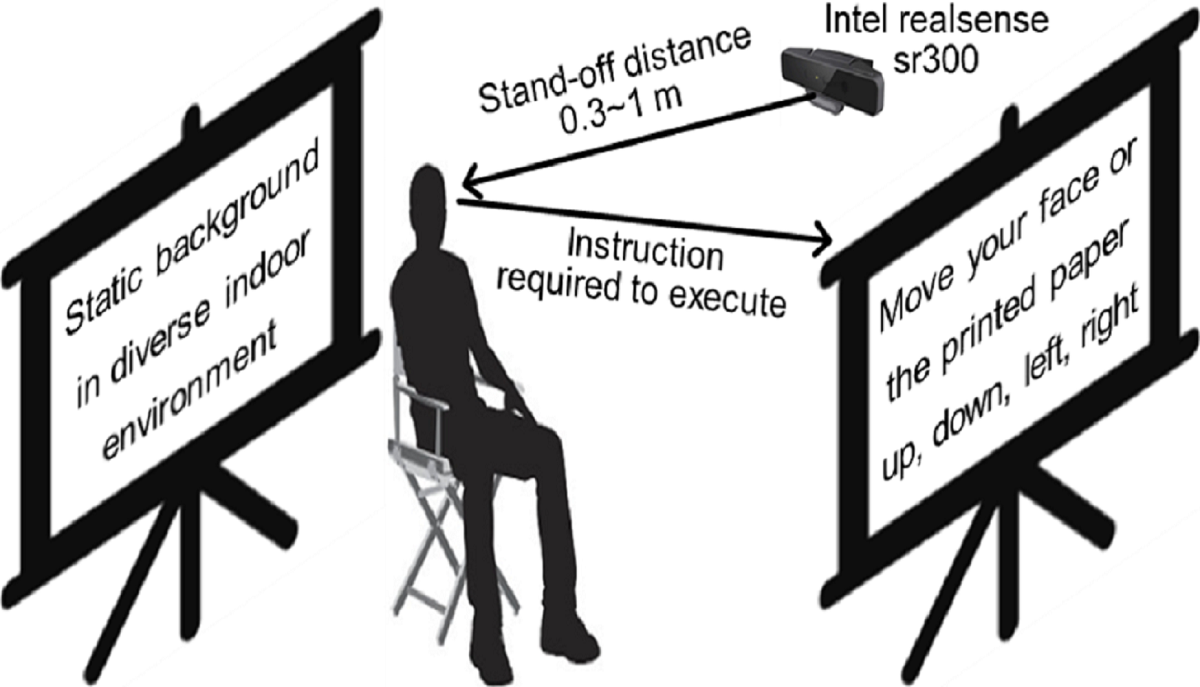}
\caption{Illustrative sketch of recording setups in the CASIA-SURF dataset.}
\label{fig:shotting}
\end{figure}

\subsection{Acquisition detail}
Fig.~\ref{fig:shotting} shows the diagram of data acquisition procedure, \ie, how the multi-modal data is recorded via the multi-modal camera in diverse indoor environment. Specifically, we use the Intel RealSense SR300 camera to capture the RGB, Depth and InfraRed (IR) videos simultaneously. During the video recording, collectors are required to turn left or right, move up or down, walk in or away from the camera. Moreover, the performers stand within the range of $0.3$ to $1.0$ meter from the camera and their face angle is asked to be less $30^0$. After that, four video streams including RGB, Depth, IR, plus RGB-Depth-IR aligned images are captured using the RealSense SDK at the same time. The resolution is $1280\times720$ for RGB images and $640\times480$ for Depth, IR and aligned images. Some examples of RGB, Depth, IR and aligned images are shown in the first column of Fig.~\ref{fig:Preprocessing}. 

\begin{figure}[t]
\centering
\includegraphics[width=1\linewidth]{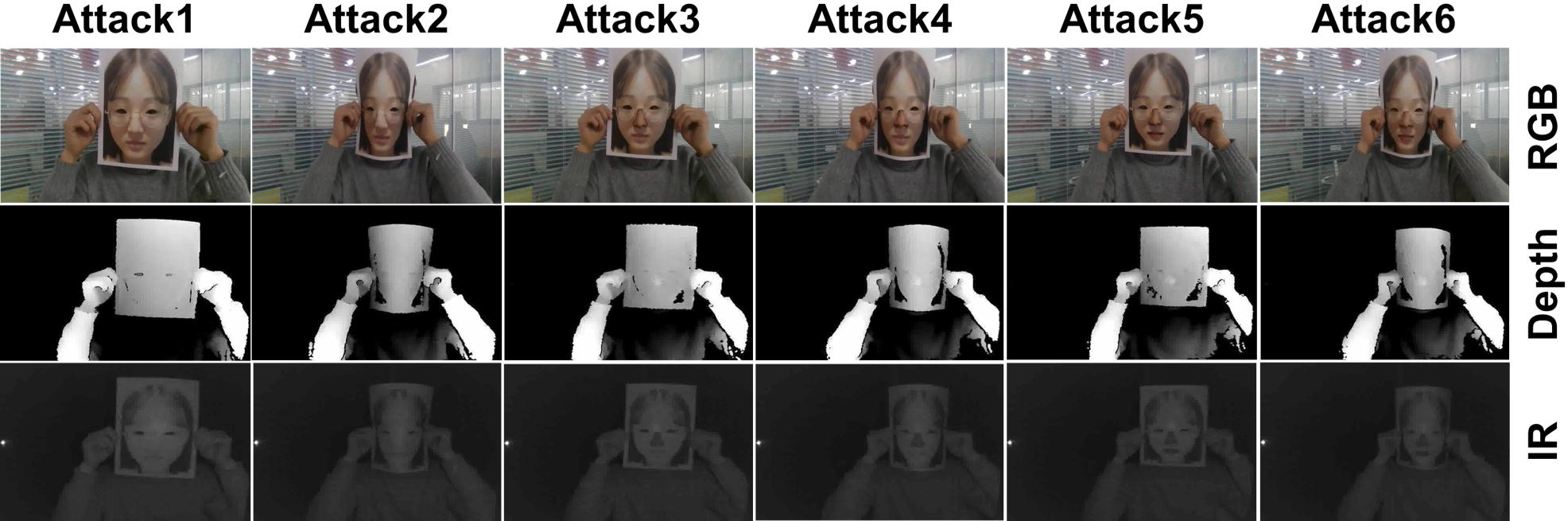}
\caption{Six attack styles in the CASIA-SURF dataset.}
\label{fig:attacks}
\end{figure}

\subsection{Attack type}
We print collectors' color pictures with A4 paper to obtain the attack faces. The printed flat or curved face images will be cut eyes, nose, mouth areas or their combinations, generating $6$ different attack ways. Thus, each sample includes $1$ live video clip and $6$ fake video clips. Fake samples are shown in Fig.~\ref{fig:attacks}. Detailed information of the $6$ attacks is given below.
\begin{itemize}
\item Attack 1: One person hold his/her flat face photo where eye regions are cut.
\item Attack 2: One person hold his/her curved face photo where eye regions are cut.
\item Attack 3: One person hold his/her flat face photo where eye and nose regions are cut.
\item Attack 4: One person hold his/her curved face photo where eye and nose regions are cut.
\item Attack 5: One person hold his/her flat face photo where eye, nose and mouth regions are cut.
\item Attack 6: One person hold his/her curved face photo where eye, nose and mouth regions are cut.
\end{itemize}

\subsection{Data preprocessing}
\begin{figure}[t]
\centering
\includegraphics[width=1\linewidth]{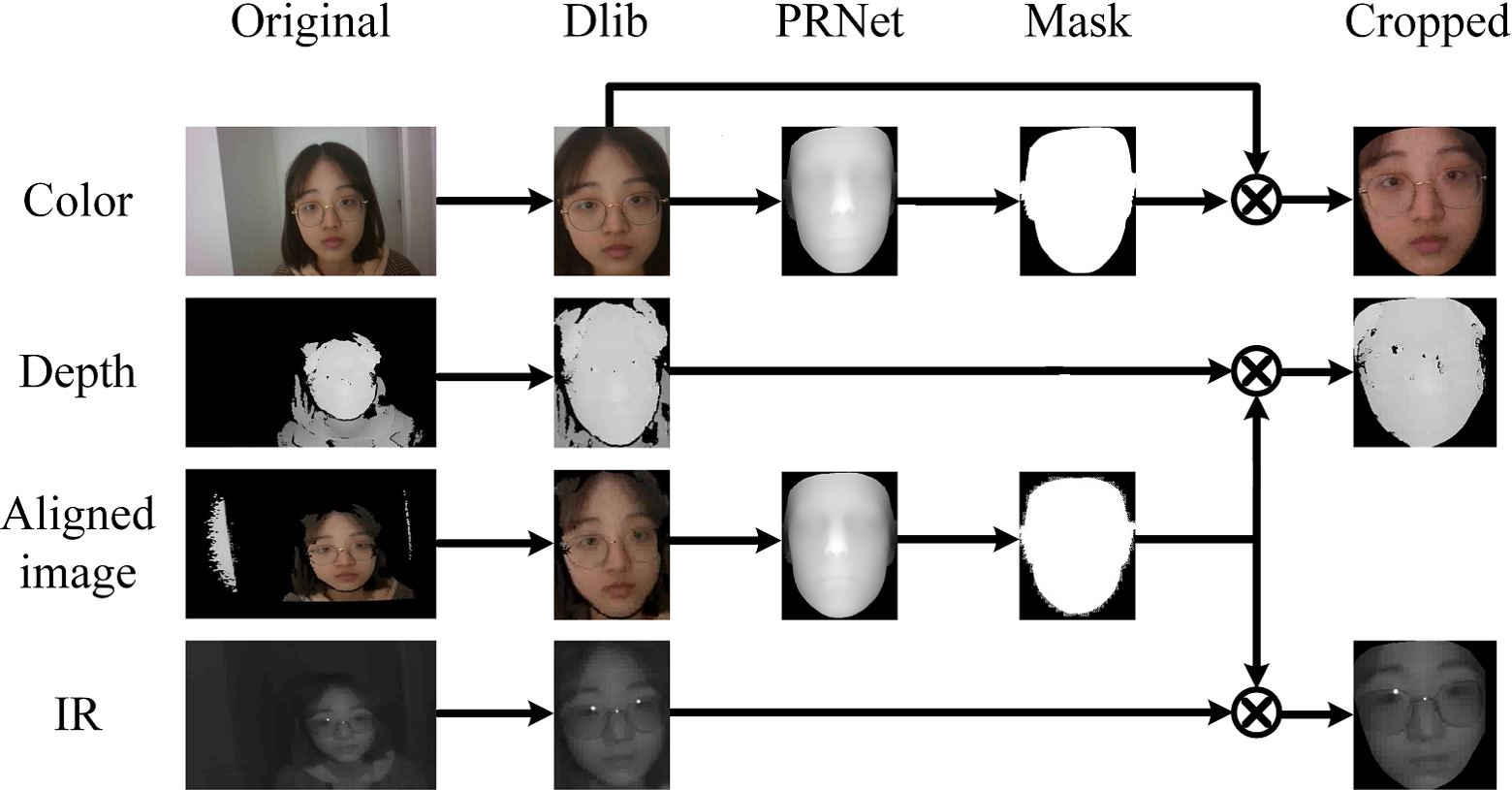}
\caption{Preprocessing details of three modalities of the CASIA-SURF dataset.}
\label{fig:Preprocessing}
\end{figure}

Data preprocessing is widely used in the face recognition system, such as face detection and face alignment. Different pre-processing methods would affect the face anti-spoofing algorithms. To focus on the face anti-spoofing task and increase the difficulty, we process the original data via face detection and alignment. As shown in Fig.~\ref{fig:Preprocessing}, we first use the Dlib~\cite{dlib09} toolkit to detect face for every frame of RGB and RGB-Depth-IR aligned videos, respectively. Then we apply the PRNet \cite{DBLP:conf/eccv/FengWSWZ18} algorithm to perform 3D reconstruction and density alignment on the detected faces. After that, we define a binary mask based on non-active face reconstruction area from previous steps. Finally, we obtain face area of RGB image via point-wise product between the RGB image and the RGB binary mask. The Depth (or IR) area can be calculated via the point-wise product between the Depth (or IR) image and the RGB-Depth-IR binary mask. After the data pre-processing stage, we manually check all the processed RGB images to ensure that they contain a high-quality large face.

\subsection{Statistics description}

Table~\ref{tab:dataset} presents the main statistics of the proposed CASIA-SURF dataset. (1) There are $1,000$ subjects with variability in terms of gender, age, glasses/no glasses and indoor environments. Each one has $1$ live video clip and $6$ fake video clips. (2) Data is divided into three subsets. The training, validation and testing subsets have $300$, $100$ and $600$ subjects with $6,300$ ($2,100$ per modality), $2,100$ ($700$ per modality), $12,600$ ($4,200$ per modality) videos, respectively. (3) From original videos, there are about $1.5$ million, $0.5$ million, $3.1$ million frames in total for training, validation, and testing subsets, respectively. Owing to the huge amount of data, we select one frame out of every $10$ frames and form the sampled set with about $151K$, $49K$, and $302K$ for training, validation and testing subsets, respectively. (4) After removing non-detected face poses with extreme lighting conditions during data prepossessing, we finally obtain about $148K$, $48K$, $295K$ images for training, validation and testing subsets in the CASIA-SURF dataset.

\begin{figure}[b]
\centering
\includegraphics[width=0.9\linewidth]{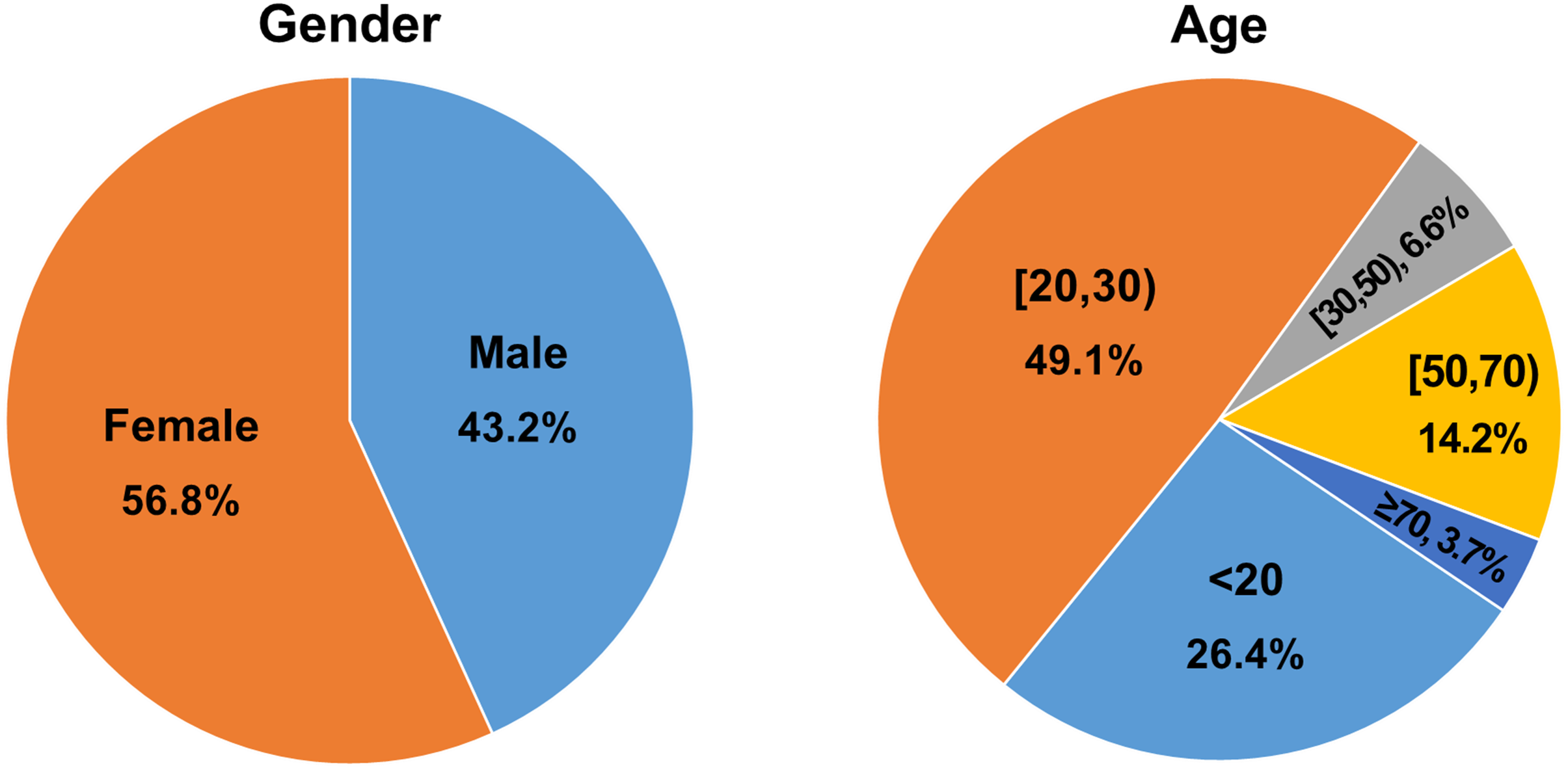}
\caption{Gender and age distribution of the CASIA-SURF dataset.}
\label{fig:statics}
\end{figure}

The information of gender statistics is shown in the left side of Fig.~\ref{fig:statics}. It shows that the ratio of female is $56.8\%$ while the ratio of male is $43.2\%$. In addition, we also show age distribution of the CASIA-SURF dataset in the right side of Fig \ref{fig:statics}. One can see a wide distribution of age ranges from $20$ to more than $70$ years old, while most of subjects are under $70$ years old. On average, the range of $[20,30)$ ages is dominant, being about $50\%$ of all the subjects.

\begin{table}[b]
\renewcommand\arraystretch{1.15}
\centering
\setlength{\tabcolsep}{3.0pt}
\caption{Statistical information of the proposed CASIA-SURF dataset.}
\small{
\begin{tabular}{|c|c|c|c|c| }
\hline
& Training & Validation & Testing & Total \\
\hline
\# Subject         & 300       & 100     & 600       & 1,000 \\
\# Video           & 6,300     & 2,100   & 12,600    & 21,000 \\
\# Original image  & 1,563,919 & 501,886 & 3,109,985 & 5,175,790 \\
\# Sampled image   & 151,635   & 49,770  & 302,559   & 503,964 \\
\# Processed image & 148,089   & 48,789  & 295,644   & 492,522 \\
\hline
\end{tabular}}
\label{tab:dataset}
\end{table}

\begin{figure*}
\centering
\subfigure[Structure of the proposed method]{
\label{fig:sef-a}
\includegraphics[width=0.61\textwidth]{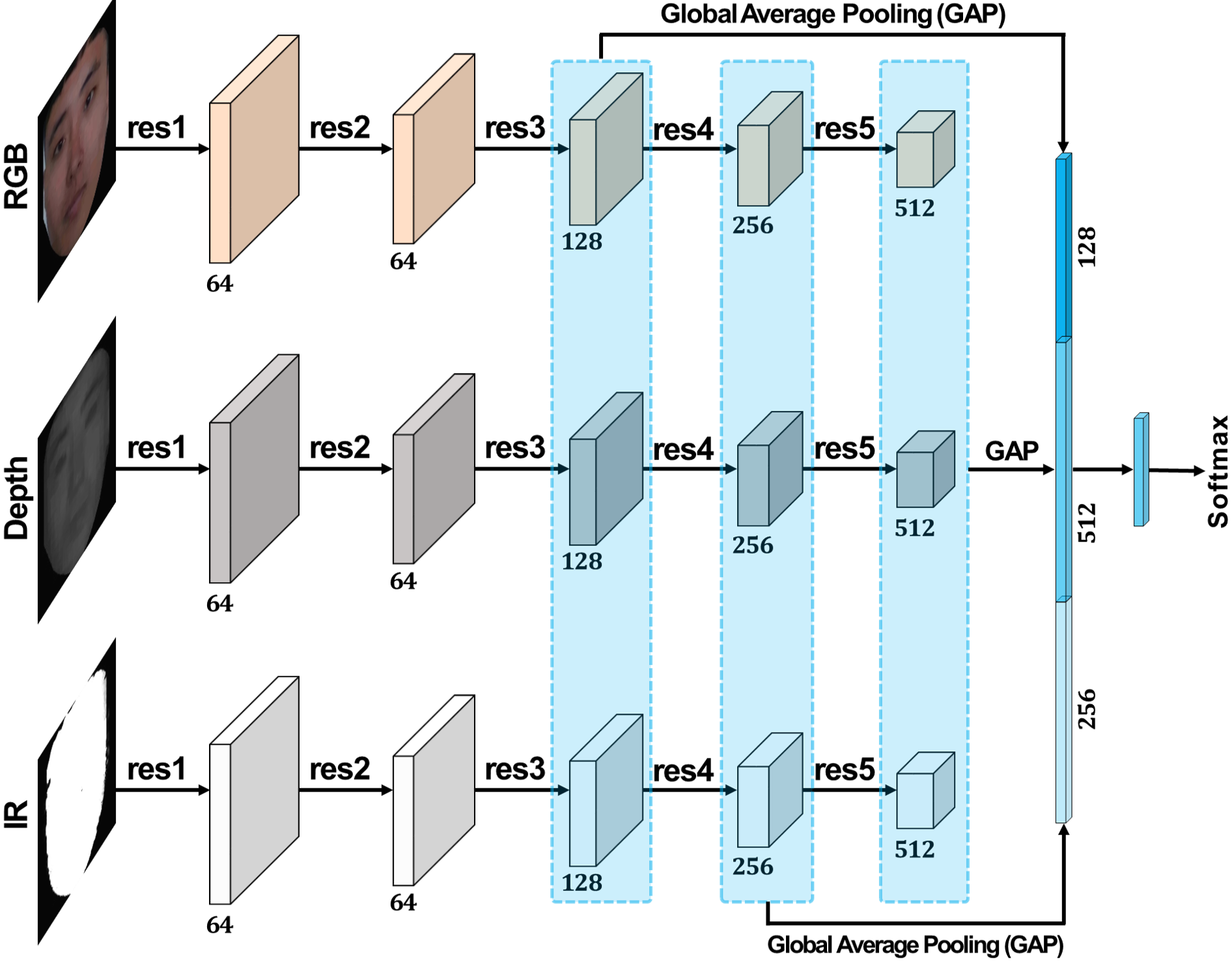}}
\ \ \ \ \ \ \ \ \ \ 
\subfigure[Squeeze-and-Excitation Fusion (SEF)]{
\label{fig:sef-b}
\includegraphics[width=0.31\textwidth]{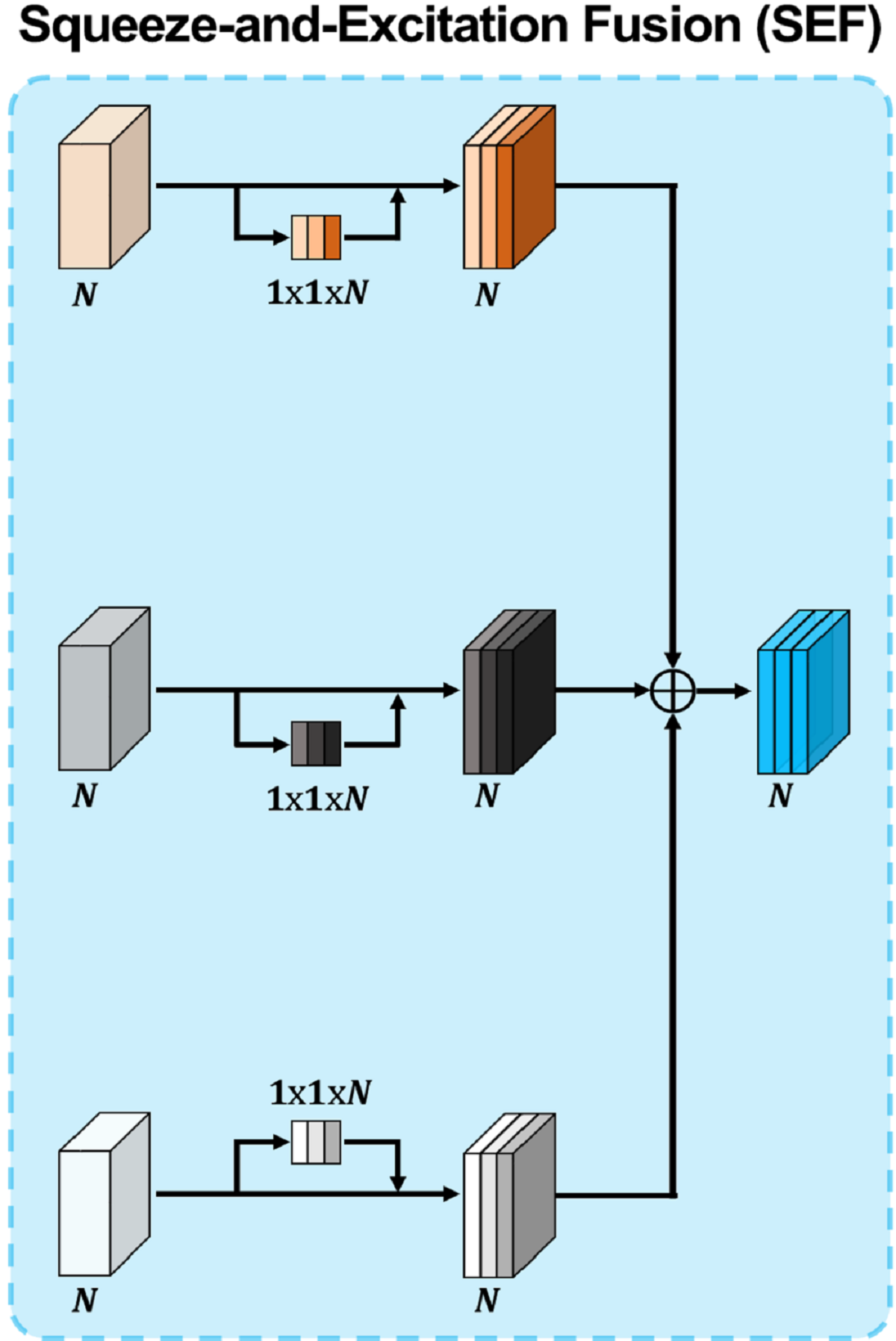}}
\caption{(a) Each stream uses ResNet-18/34 as backbone, which has five convolution blocks (\ie, res1, res2, res3, res4, res5) to extract features of each modal data (\ie, RGB, Depth, IR). We first fuse features from different modalities via SEF after res3, res4 and res5 respectively, then squeeze these fused features via GAP, next concatenate these squeezed features and finally use the concatenated features to predict real and fake. (b) Illustration of SEF.}
\label{fig:sef}
\end{figure*}

\subsection{Evaluation protocol}
We select the live faces and Attacks 4, 5, 6 as the training subset, while the live faces and Attacks 1, 2, 3 as the validation and testing subsets. This makes the ratio of flat/curved face and the extend of cut organ different between training and evaluation in order to increase the difficulty. The validation subset is used for model and hyper-parameter selection and the testing subset for final evaluation. Our dataset has two types of evaluation protocol: (1) \textbf{within-modal evaluation}, in which algorithms are trained and evaluated in the same modalities; (2) \textbf{cross-modal evaluation}, in which algorithms are trained in one modality while evaluated in other modalities.

\subsection{Evaluation metric}
Following the face recognition task, we use the ROC curve as the main evaluation metric. ROC curve is a suitable indicator for the algorithms applied in the real world applications, because we can select a suitable trade-off threshold between FPR and TPR according to the requirements. Empirically, we compute TPR@FPR=$10^{-2}$, $10^{-3}$ and $10^{-4}$ as the quantitative indicators. Among them, we regard TPR@FPR=$10^{-4}$ as the main comparison. Besides, the commonly used metric ACER, APCER and NPCER are also provided for reference.

\begin{table*}
\renewcommand\arraystretch{1.2}
\centering
\setlength{\tabcolsep}{11pt}
\caption{Effectiveness of the proposed method. All ResNet-18 models are trained on training set and evaluated on testing set. The testing set has $57,710$ triples (\ie, RGB, Depth, IR images), composed by $17,458$ real face triplets and $40,252$ fake face triplets.}
\small{
\begin{tabular}{|c|c|c|c|c|c|c|}
\hline
\multirow{3}{*}{Method} & \multicolumn{3}{c|}{TPR (\%) / \# TP} & \multirow{3}{*}{APCER (\%)} & \multirow{3}{*}{NPCER (\%)} & \multirow{3}{*}{ACER (\%)}\\
\cline{2-4}
 & @FPR=$10^{-2}$ &@FPR=$10^{-3}$ &@FPR=$10^{-4}$ & &  & \\
 & (\# FP$\approx$403) & (\# FP$\approx$40) & (\# FP$\approx$4) & &  & \\
\hline
\hline
Halfway fusion &89.1 / 15,555 &33.6 / 5,866 &17.8 / 3,108 &5.6 &3.8 &4.7 \\
\hline
\hline
SEF &96.7 / 16,882 &81.8 / 14,281 &56.8 / 9,916 &3.8 &1.0 &2.4\\
\hline
+ Data augmentation &97.8 / 17,074 &84.8 / 14,804 &66.2 / 11,557 &3.7 &0.5 &2.1 \\
\hline
+ Addition operation &98.7 / 17,231 &93.2 / 16,271 &73.5 / 12,832 &2.8 &0.3 &1.5 \\
\hline
+ ImageNet pretrain &99.4 / 17,353 &95.8 / 16,725 &81.4 / 14,211 &2.3 &0.3 &1.3 \\
\hline
+ Multi-scale fusion &99.7 / 17,406 &97.4 / 17,004 &92.4 / 16,131 &1.9 &0.1 &1.0 \\
\hline
+ Stronger backbone &\textbf{99.8 / 17,423} &\textbf{98.4 / 17,179} &\textbf{95.2 / 16,620} &\textbf{1.6} &\textbf{0.08} &\textbf{0.8}\\					
\hline
\end{tabular}}
\label{tab:ablation}
\end{table*}

\section{Proposed method}
Before showing some experimental analysis on the proposed dataset, we first built a strong baseline. We aim to find a straightforward architecture that provides good performance on our CASIA-SURF. Thus, we regard the face anti-spoofing problem as a binary classification task (fake \emph{v.s} real) and conduct the experiments based on the ResNet-18/34 \cite{he2016deep} classification network. ResNet-18/34 consist of five convolutional blocks (namely res1, res2, res3, res4, res5), a global average pooling layer and a softmax layer, which are relatively shallow networks with high classification performance.

\subsection{Naive halfway fusion}
CASIA-SURF is characterized by multi-modality (\ie, RGB, Depth, IR) and a key issue is how to fuse the complementary information between the three modalities. We use a multi-stream architecture with three subnetworks to study the dataset modalities, in which RGB, Depth and IR data are learnt separately by each stream, and then shared layers are appended at a point to learn joint representations and perform cooperated decisions. The halfway fusion is one of the commonly used fusion methods, which combines the subnetworks of different modalities at a later stage, \ie, immediately after the third convolutional block (res3) via the feature map concatenation. In this way, features from different modalities can be fused to perform classification. However, direct concatenating these features cannot make full use of the characteristics between different modalities.

\subsection{Squeeze and excitation fusion}
The three modalities provide with complementary information for different kind of attacks: RGB data have rich appearance details, Depth data are sensitive to the distance between the image plane and the corresponding face, and IR data measure the amount of heat radiated from a face. Inspired by~\cite{hu2018senet}, we propose the Squeeze and Excitation Fusion (SEF) module to fuse features from different modalities. As shown in Fig.~\ref{fig:sef-b}, this module first adds a branch\footnote{It is the same as the ``Squeeze-and-Excitation'' branch~\cite{hu2018senet}, composed of one global average pooling layer and two consecutive fully connected layers} to obtain the channel-wise weights for each modality, then re-weights the input features and finally combines these re-weighted features together. Comparing to the naive halfway fusion that directly combines the features from different modalities, the SEF performs modality-dependent feature re-weighting to select the more informative channel features while suppressing less useful features from each modality.

\subsection{From single-scale to multi-scale SEF}
In our previous work~\cite{DBLP:conf/cvpr/abs-1812-00408}, we only apply the SEF module on one of the scales in the ResNet-18 network, \ie, the SEF module is appended after the res3 block to fuse features from different modalities and the subsequent blocks are shared. As is well-known, in convolutional neural networks, the high-level layer has a large receptive field with strong ability to represent semantic information, but has low resolution with weak ability to represent detailed information. While the low-level layer has a small receptive field with weak ability to represent semantic information, but has large resolution with strong ability to represent detailed information. For the anti-spoofing task, it is better to fuse deep features with strong semantic information and shallow features with detailed information to globally and locally determine whether a face is real or fake.

However, the single-scale SEF does not make full use of features from different levels. To this end, we extend the SEF from single scale to multiple scales. As shown in Fig.~\ref{fig:sef-a}, our proposed method has a three-stream architecture and each subnetwork is feed with the image of different modalities. The res1, res2, res3, res4 and res5 blocks from each stream extract features from different modalities. After that, we first fuse features from different modalities via the SEF after res3, res4 and res5 respectively, then squeeze these fused features via the Global Average Pooling (GAP), next concatenate these squeezed features and finally use the concatenated features to predict real and fake.

\begin{table*}
\renewcommand\arraystretch{1.2}
\centering
\setlength{\tabcolsep}{12pt}
\caption{Effect of number of modalities. All models are based on ResNet-18 and trained on the CASIA-SURF training subset and tested on the testing subset with one, or two, or three modalities.}
\small{
\begin{tabular}{|c|c|c|c|c|c|c|c|c|}
\hline
\multirow{2}{*}{Modality} & \multicolumn{3}{c|}{TPR (\%)} & \multirow{2}{*}{APCER (\%)} & \multirow{2}{*}{NPCER (\%)} & \multirow{2}{*}{ACER (\%)}\\
\cline{2-4}
 & @FPR=$10^{-2}$ &@FPR=$10^{-3}$ &@FPR=$10^{-4}$ & &  & \\
\hline
\hline
RGB        &51.7 &27.5 &14.6 &40.3 &1.6 &21.0 \\
\hline
Depth      &96.8 &86.5 &67.3 &6.0  &1.2 &3.6 \\
\hline
IR         &62.5 &29.4 &15.9 &38.6 &0.4 &19.4 \\
\hline
RGB\&Depth &97.1 &87.5 &71.1 &5.8 &0.8 &3.3 \\
\hline
RGB\&IR    &87.4 &60.3 &37.0 &36.5 &0.005 &18.3 \\
\hline
Depth\&IR  &99.4 &95.2 &81.2 &2.0  &0.3 &1.1 \\
\hline
RGB\&Depth\&IR &99.7 &97.4 &92.4 &1.9 &0.1 &1.0 \\
\hline
\end{tabular}}
\label{tab:modalities}
\end{table*}

\begin{figure*}[t]
\centering
\subfigure[]{
\label{fig:identity}
\includegraphics[width=0.4\textwidth]{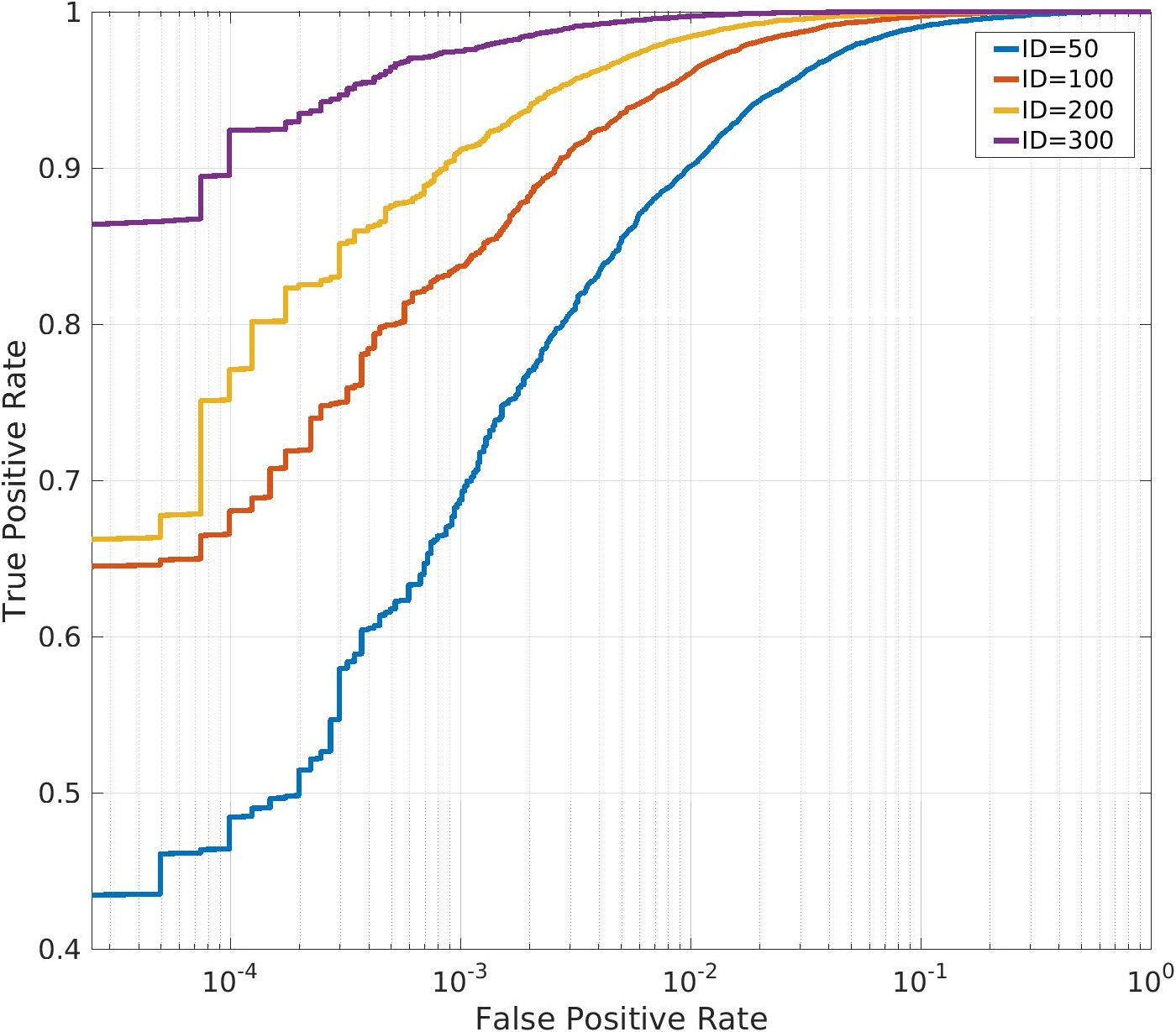}}
\ \ \ \ \ \ \ \ \ \ \ \ 
\subfigure[]{
\label{fig:acer}
\includegraphics[width=0.4\textwidth]{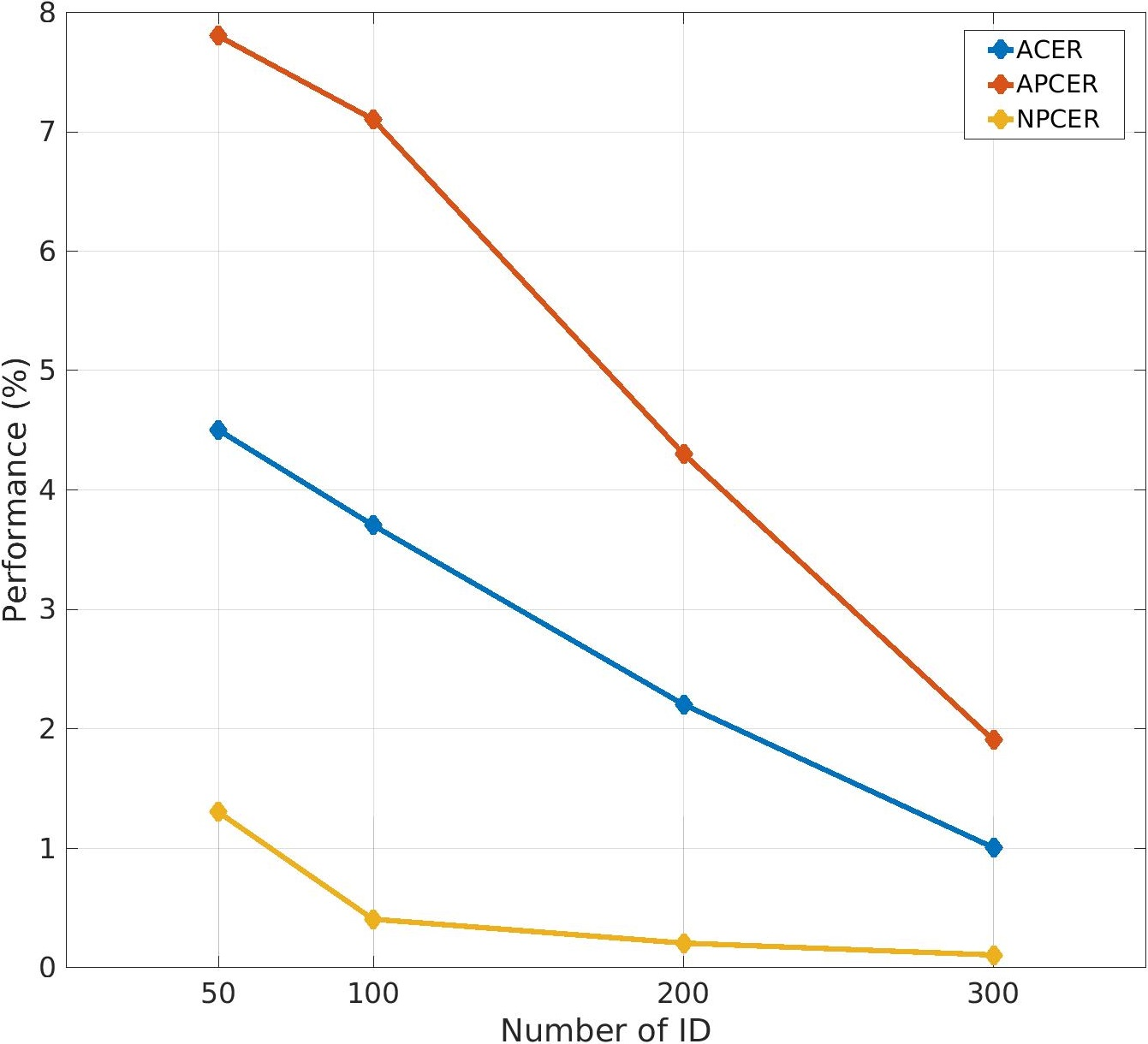}}
\caption{(a) ROC curves of different training subset size in the CASIA-SURF dataset. (b) Performance \emph{vs.} training subset size in the CASIA-SURF dataset.}
\vspace{-1mm}
\end{figure*}

\section{Experiments}
In this section, we firstly describe the implementation details, secondly verify the effectiveness of the proposed method, thirdly present a series of experiments to analyze the CASIA-SURF dataset in terms of number of modalities and subjects, fourthly conduct the cross-modal evaluation and finally present the generalization capability of the proposed dataset.

\subsection{Implementation detail}
We resize the cropped face region to $112\times112$, and use random flipping, rotation, resizing, cropping and color distortion for data augmentation. For the CASIA-SURF dataset analyses, all models are trained for $40$ epochs and the initial learning rate is $0.01$, decreased by a factor of $10$ after $20$ and $30$ epochs, respectively. All models are optimized via the Adaptive Moment Estimation (Adam) algorithm on $2$ TITAN X (Maxwell) GPU with a mini-batch $256$. Weight decay and momentum are set to $0.0005$ and $0.9$, respectively.

\subsection{Model analysis}
As listed in Table~\ref{tab:ablation}, we carry out some ablation experiments on the CASIA-SURF dataset to analyze our proposed method. For a fair comparison, we use the same settings except for the specific modification. In the conference version of this work~\cite{DBLP:conf/cvpr/abs-1812-00408}, we have verified the effectiveness of the single-scale SEF module, which improves the TPR@FPR=[$10^{-2}$, $10^{-3}$, $10^{-4}$], APCER, NPCER, ACER from $89.1\%$, $33.6\%$, $17.8\%$, $5.6\%$, $3.8\%$, $4.7\%$ to $96.7\%$, $81.8\%$, $56.8\%$, $3.8\%$, $1.0\%$, $2.4\%$, respectively. At this stage, the commonly used metrics APCER, NPCER and ACER are very promising, but TPR@FPR=[$10^{-2}$, $10^{-3}$, $10^{-4}$] have a big space to improve, especially for TPR@FPR=$10^{-4}$. To this end, we explore some strategies as shown in Table~\ref{tab:ablation} to further improve the performance: (1) adjusting some hyper-parameters of data augmentation increases TPR by $1.1\%$, $3.0\%$, $9.4\%$ for FPR=$10^{-2}$, $10^{-3}$, $10^{-4}$; (2) replacing the concatenation operation in the SEF module with the addition operation boosts TPR@FPR=[$10^{-2}$, $10^{-3}$, $10^{-4}$] by $0.9\%$, $8.4\%$, $7.3\%$; (3) using ImageNet pretrained model brings $0.7\%$, $2.6\%$, $7.9\%$ improvements for TPR@FPR=[$10^{-2}$, $10^{-3}$, $10^{-4}$]; (4) extending the SEF from single scale to multiple scales improves TPR@FPR=[$10^{-2}$, $10^{-3}$, $10^{-4}$] to $99.7\%$, $97.4\%$, $92.4\%$; (5) applying a stronger backbone from ResNet-18 to ResNet-34 has $0.1\%$, $1.0\%$, $2.8\%$ improvements for TPR@FPR=[$10^{-2}$, $10^{-3}$, $10^{-4}$]. Besides, the APCER, NPCER and ACER are also improved from $3.8\%$, $1.0\%$, $2.4\%$ to $1.6\%$, $0.08\%$, $0.8\%$ after using these new strategies. Notably, the newly proposed multi-scale SEF achieves the most significant improvement $11.0\%$ for TPR@FPR=$10^{-4}$, demonstrating its effectiveness.

\subsection{Dataset analysis}
The proposed CASIA-SURF dataset has three modalities with $1,000$ subjects. In this subsection, we analyze the effect of the number of modalities and subjects.

{\flushleft \textbf{Effect of number of modalities. }}
As shown in Table~\ref{tab:modalities}, only using the prevailing RGB data, the results are $51.7\%$, $27.5\%$, $14.6\%$ for TPR@FPR=[$10^{-2}$, $10^{-3}$, $10^{-4}$] and $40.3\%$, $1.6\%$, $21.0\%$ for APCER, NPCER, ACER. In contrast, simply using the IR data, the results can be improved to $62.5\%$ (TPR@FPR=$10^{-2}$), $29.4\%$ (TPR@FPR=$10^{-3}$), $15.9\%$ (TPR@FPR=$10^{-4}$), $38.6\%$ (APCER), $0.4\%$ (NPCER) and $19.4\%$ (ACER), respectively. Among these three modalities, the Depth data achieves the best performance, \ie, $96.8\%$, $86.5\%$, $67.3\%$ for TPR@FPR=[$10^{-2}$, $10^{-3}$, $10^{-4}$], $6.0\%$ for APCER and $3.6\%$ for ACER. By fusing the data of arbitrary two modalities or all three ones, we observe an increase in performance. The best results are achieved by fusing all the three modalities, improving the best results of single modality from $96.8\%$, $86.5\%$, $67.3\%$, $6.0\%$, $0.4\%$, $3.6\%$ to $99.7\%$, $97.4\%$, $92.4\%$, $1.9\%$, $0.1\%$, $1.0\%$ for TPR@FPR=[$10^{-2}$, $10^{-3}$, $10^{-4}$], APCER, NPCER, ACER, respectively, demonstrating the necessity of multi-modal dataset.

\begin{table*}
\renewcommand\arraystretch{1.2}
\centering
\setlength{\tabcolsep}{10.8pt}
\caption{Cross-modal evaluation. All models are based on ResNet-18 and trained on CASIA-SURF training set and tested on testing set.}
\small{
\begin{tabular}{|c|c|c|c|c|c|c|c|c|c|}
\hline
\multicolumn{2}{|c|}{Modality} & \multicolumn{3}{c|}{TPR (\%)} & \multirow{2}{*}{APCER (\%)} & \multirow{2}{*}{NPCER (\%)} & \multirow{2}{*}{ACER (\%)}\\
\cline{1-5} Training & Testing &@FPR=$10^{-2}$ &@FPR=$10^{-3}$ &@FPR=$10^{-4}$ & &  & \\
\hline
\hline
RGB   & Depth &16.8 &1.6 &0.1 &82.9 &0.8 &41.8 \\
\hline
RGB   & IR    &4.0 &0.2 &0.02 &73.8 &0.4 &37.1\\
\hline
Depth & RGB   &6.9 &2.1 &0.7 &42.4 &38.6 &40.5 \\
\hline
Depth & IR    &6.0 &1.4 &0.3 &3.7 &86.5 &45.1 \\
\hline
IR    & RGB   &4.4 &0.4 &0.04 &93.9 &4.9 &49.4 \\
\hline
IR    & Depth &0.09 &0.01 &0.001 &60.2 &95.9 &78.1\\
\hline
\end{tabular}}
\label{tab:cross-modal}
\end{table*}

{\flushleft \textbf{Effect of number of subjects. }}
As described in~\cite{DBLP:conf/iccv/SunSSG17}, there is a logarithmic relation between the amount of training data and the performance. To quantify the impact of having a large amount of training data in PAD, we show how the performance grows as training data increases in our benchmark. For this purpose, we train our baselines with different sized subsets of subjects randomly sampled from the training subset. This is, we randomly select $50$, $100$ and $200$ from $300$ subjects for training. Fig.~\ref{fig:identity} shows ROC curves for different number of subjects. We can see that the TPR is better when more subjects are used for training across different FPR. When FPR=$10^{-4}$, the best TPR of $300$ subjects is higher about $15\%$ than the second best TPR result (ID=$200$), showing that using more data will achieve better performance. In Fig.~\ref{fig:acer}, we also provide with the performance of ACER, APCER and NPCER under different number of subjects. Their performances are getting better when more subjects are considered.

{\flushleft \textbf{Difficulty of different types of attacks. }}
Since the Attacks 1, 2, 3 are selected as the validation and testing subsets, thus we evaluate the trained model (\ie, TPR@FPR=$10^{-4}$ is $92.4\%$) on Attack 1, Attack 2 and Attack 3, respectively. The corresponding TPR@FPR=$10^{-4}$ performances are $94.4\%$, $92.9\%$ and $86.4\%$. The difference between Attack 1 and Attack 2 is whether the fake face is curved. Attack 2 is more challenging than Attack 1, indicating that the curved fake face is more difficult than the flat fake face. The difference between Attack 1 and Attack 3 is whether the fake face is cutout. Attack 3 is more challenging than Attack 1, indicating that the cutout fake face is more difficult than the intact fake face.

\subsection{Cross-modal evaluation}
We introduce the cross-modality evaluation protocol for the academic community to explore new issues. Although there are no real world scenarios for this protocol until now, if algorithms trained on a certain modality data are able to perform well on other modalities data, this will greatly enhance their versatility for different scenes with different devices. We aim to provide this cross-modal evaluation protocol for those possible real-world scenarios in the future. In this protocol, one of RGB, Depth and IR modalities is used for training, and then evaluate on the remaining modalities. As shown in Table~\ref{tab:cross-modal}, the model only trained on the RGB, Depth or IR modality is evaluated on the Depth and IR, RGB and IR, RGB and Depth modalities, respectively. All the results are far away from satisfactory, even worse than random guesses. The reason behind these poor results is the large differences between different modalities data. Therefore, it is a challenging task and deserves further study in academic community.

\subsection{Using CASIA-SURF for pre-training}
The CASIA-SURF dataset contains not only RGB images, but also the corresponding Depth information, which is indeed beneficial for Depth supervised face anti-spoofing methods~\cite{Liu2018Learning,FASTD2018arxiv}. Thus, we adopt FAS-TD-SF~\cite{FASTD2018arxiv} as our baseline to evaluate the generalization capability of the proposed dataset. We first pre-train the model on CASIA-SURF and then fine-tune with the concerned dataset including Oulu-NPU~\cite{Boulkenafet2017OULU}, SiW~\cite{Liu2018Learning} and CASIA-MFSD~\cite{Zhang2012A}. This
model is termed as FAS-TD-SF (CASIA-SURF).

\begin{table}[h]
\renewcommand\arraystretch{1.15}
\centering
\setlength{\tabcolsep}{2pt}
\caption{Evaluation results on four protocols of Oulu-NPU.}
\footnotesize{
\begin{tabular}{|c|c|c|c|c|}
\hline
Prot. & Method & APCER (\%) & NPCER (\%) & ACER (\%) \\
\hline
\hline
\multirow{6}{*}{1} & CPqD~\cite{Boulkenafet2017A} & 2.9 & 10.8 & 6.9 \\
\cline{2-5} & GRADIANT~\cite{Boulkenafet2017A} & 1.3 & 12.5 & 6.9 \\
\cline{2-5} & FAS-BAS~\cite{Liu2018Learning} & 1.6 & \textbf{1.6} & 1.6 \\
\cline{2-5} & FAS-Ds~\cite{Jourabloo2018Face} & 1.2 & 1.7 & \textbf{1.5} \\
\cline{2-5} & STASN~\cite{yang2019face} & 1.2 & 2.5 & 1.9 \\
\cline{2-5} & FAS-TD-SF~\cite{FASTD2018arxiv} & \textbf{0.8} & 10.8 & 5.8 \\
\cline{2-5} & FAS-TD-SF (CASIA-SURF) & 2.7 & 2.5 & 2.6 \\
\hline
\hline
\multirow{6}{*}{2} & MixedFASNet~\cite{Boulkenafet2017A} & 9.7 & 2.5 & 6.1 \\
\cline{2-5} & FAS-Ds~\cite{Jourabloo2018Face} & 4.2 & 4.4 & 4.3 \\
\cline{2-5} & FAS-BAS~\cite{Liu2018Learning} & \textbf{2.7} & 2.7 & 2.7 \\
\cline{2-5} & GRADIANT~\cite{Boulkenafet2017A} & 3.1 & 1.9 & 2.5 \\
\cline{2-5} & STASN~\cite{yang2019face} & 4.2 & \textbf{0.3} & \textbf{2.2} \\
\cline{2-5} & FAS-TD-SF~\cite{FASTD2018arxiv} & 3.6 & 3.8 & 3.7 \\
\cline{2-5} & FAS-TD-SF (CASIA-SURF) & \textbf{2.7} & 1.6 & \textbf{2.2} \\
\hline
\hline
\multirow{6}{*}{3} & MixedFASNet~\cite{Boulkenafet2017A} & 5.3$\pm$6.7 & 7.8$\pm$5.5 & 6.5$\pm$4.6 \\
\cline{2-5} & GRADIANT~\cite{Boulkenafet2017A} & 2.6$\pm$3.9 & 5.0$\pm$5.3 & 3.8$\pm$2.4 \\
\cline{2-5} & FAS-Ds~\cite{Jourabloo2018Face} & 4.0$\pm$1.8 & 3.8$\pm$1.2 & 3.6$\pm$1.6 \\
\cline{2-5} & FAS-BAS~\cite{Liu2018Learning} & 2.7$\pm$1.3 & 3.1$\pm$1.7 & 2.9$\pm$1.5 \\
\cline{2-5} & STASN~\cite{yang2019face} & 4.7$\pm$3.9 & \textbf{0.9$\pm$1.2} & 2.8$\pm$1.6 \\
\cline{2-5} & FAS-TD-SF~\cite{FASTD2018arxiv} & 3.1$\pm$1.8 & 6.6$\pm$9.4 & 5.3$\pm$4.4 \\
\cline{2-5} & FAS-TD-SF (CASIA-SURF) & \textbf{2.4$\pm$1.5} & 2.2$\pm$3.8 & \textbf{2.3$\pm$2.6} \\
\hline
\hline
\multirow{6}{*}{4} & Massy\_HNU~\cite{Boulkenafet2017A} & 35.8$\pm$35.3 & 8.3$\pm$4.1 & 22.1$\pm$17.6 \\
\cline{2-5} & GRADIANT~\cite{Boulkenafet2017A} & \textbf{5.0$\pm$4.5} & 15.0$\pm$7.1 & 10.0$\pm$5.0 \\
\cline{2-5} & FAS-BAS~\cite{Liu2018Learning} & 9.3$\pm$5.6 & 10.4$\pm$6.0 & 9.5$\pm$6.0 \\
\cline{2-5} & FAS-Ds~\cite{Jourabloo2018Face} & 5.1$\pm$6.3 & 6.1$\pm$5.1 & \textbf{5.6$\pm$5.7} \\
\cline{2-5} & STASN~\cite{yang2019face} & 6.7$\pm$10.6 & 8.3$\pm$8.4 & 7.5$\pm$4.7 \\
\cline{2-5} & FAS-TD-SF~\cite{FASTD2018arxiv} & 7.0$\pm$5.3 & 20.0$\pm$24.8 & 13.5$\pm$10.9 \\
\cline{2-5} & FAS-TD-SF (CASIA-SURF) & 8.7$\pm$5.6 & \textbf{5.8$\pm$8.0} & 7.2$\pm$5.8 \\
\hline
\end{tabular}
}
\label{tab:Oulu}
\end{table}

{\noindent \textbf{Oulu-NPU dataset.}} It is a high-resolution dataset, consisting of $4,950$ real access and spoofing videos with many real-world variations. This dataset contains $4$ evaluation protocols to validate the generalization of methods: Protocol 1 evaluates on the illumination variation; Protocol 2 examines the influence of different attack medium, such as unseen printers or displays; Protocol 3 studies the effect of the input camera variation; Protocol 4 considers all the factors above, which is the most challenging. As shown in Table~\ref{tab:Oulu}, using the proposed dataset to pre-train our baseline method FAS-TD-SF significantly improves its ACER performance, \ie, from $5.8\%$ to $2.6\%$ in Protocol 1, from $3.7\%$ to $2.2\%$ in Protocol 2, from $5.3\%$ to $2.3\%$ in Protocol 3, and from $13.5\%$ to $7.2\%$ in Protocol 4. Without bells and whistles, our method achieves the lowest ACER in $2$ out of $3$ protocols. We believe that other state-of-the-art methods can be further improved by using our CASIA-SURF as the pre-training dataset.

{\noindent \textbf{SiW dataset.}} It contains more live subjects and has three protocols used for evaluation, please refer to~\cite{Liu2018Learning} for more details of the protocols. Table~\ref{tab:SiW} shows the comparison between three state-of-the-art methods on the SiW dataset. FAS-TD-SF generally achieves better performance than FAS-BAS, while our pre-trained FAS-TD-SF on CASIA-SURF can further improve the performance across all protocols. Concretely, the performance of ACER is reduced by $0.25\%$, $0.14\%$ and $1.38\%$ in Protocol 1, 2, and 3 respectively when using the proposed CASIA-SURF dataset as pre-training. The improvement indicates that pre-training on the proposed dataset supports the generalization on data containing variabilities in terms of (1) face pose and expression, (2) replay attack mediums, and (3) cross Presentation Attack Instruments (PAIs), such as from print attack to replay attack. Interestingly, it also demonstrates our dataset is also useful to be used for pre-trained models when replay attack mediums cross PAIs.

\begin{table}[h]
\renewcommand\arraystretch{1.15}
\centering
\setlength{\tabcolsep}{2.6pt}
\caption{Evaluation results on three protocols of SiW. }
\footnotesize{
\begin{tabular}{|c|c|c|c|c|}
\hline
Prot. & Method & APCER(\%) & NPCER(\%) & ACER(\%) \\
\hline
\hline
\multirow{3}{*}{1} &FAS-BAS~\cite{Liu2018Learning} &3.58 &3.58 &3.58 \\
\cline{2-5} &STASN~\cite{yang2019face} &- &- &1.00 \\
\cline{2-5} &FAS-TD-SF~\cite{FASTD2018arxiv} &\textbf{1.27} &0.83 &1.05 \\
\cline{2-5} &FAS-TD-SF (CASIA-SURF) &\textbf{1.27} &\textbf{0.33} &\textbf{0.80} \\
\hline
\multirow{3}{*}{2} &FAS-BAS~\cite{Liu2018Learning} &0.57$\pm$0.69 &0.57$\pm$0.69 &0.57$\pm$0.69 \\
\cline{2-5} &STASN~\cite{yang2019face} &- &- &0.28$\pm$0.05 \\
\cline{2-5} &FAS-TD-SF~\cite{FASTD2018arxiv} &0.33$\pm$0.27 &0.29$\pm$0.39 &0.31$\pm$0.28 \\
\cline{2-5} &FAS-TD-SF (CASIA-SURF) &\textbf{0.08$\pm$0.17} &\textbf{0.25$\pm$0.22} &\textbf{0.17$\pm$0.16} \\
\hline
\multirow{3}{*}{3} &FAS-BAS~\cite{Liu2018Learning} &8.31$\pm$3.81 &8.31$\pm$3.80 &8.31$\pm$3.81 \\
\cline{2-5} &STASN~\cite{yang2019face} &- &- &12.10$\pm$1.50 \\
\cline{2-5} &FAS-TD-SF~\cite{FASTD2018arxiv} &7.70$\pm$3.88 &7.76$\pm$4.09 &7.73$\pm$3.99 \\
\cline{2-5} &FAS-TD-SF (CASIA-SURF) &\textbf{6.27$\pm$4.36} &\textbf{6.43$\pm$4.42} &\textbf{6.35$\pm$4.39} \\
\hline
\end{tabular}
}
\label{tab:SiW}
\end{table}

{\noindent \textbf{CASIA-MFSD dataset.}} It contain low-resolution videos with resolution $640\times480$ and $1280\times720$. To further evaluate the generalization capability of the proposed dataset, we perform cross-testing experiments on this dataset, \ie, training on the proposed CASIA-SURF and then directly evaluating on the CASIA-MFSD dataset. State-of-the-art methods~\cite{Pereira2013Can, Bharadwaj2013Computationally, pinto2015face, yang2014learn} are listed for comparison, which use the Replay-Attack~\cite{Chingovska_BIOSIG-2012} dataset for training. Results in Table~\ref{tab:CASIA-MFSD} show that the model trained on the CASIA-SURF dataset performs the best among all models.

\begin{table}[h]
\renewcommand\arraystretch{1.15}
\centering
\setlength{\tabcolsep}{6pt}
\caption{Evaluation results on different cross-testing protocols.}
\footnotesize{
\begin{tabular}{|c|c|c|c|}
\hline
Method & Training & Testing & HTER (\%)\\
\hline
\hline
Motion~\cite{Pereira2013Can} &Replay-Attack &CASIA-MFSD &47.9 \\
\hline
LBP~\cite{Pereira2013Can} &Replay-Attack &CASIA-MFSD &57.6 \\
\hline
Motion-Mag~\cite{Bharadwaj2013Computationally} &Replay-Attack &CASIA-MFSD &47.0 \\
\hline
Spectral cubes~\cite{pinto2015face} &Replay-Attack &CASIA-MFSD &50.0\\
\hline
CNN~\cite{yang2014learn} &Replay-Attack &CASIA-MFSD &45.5\\
\hline
STASN~\cite{yang2019face} &Replay-Attack &CASIA-MFSD &\textbf{30.9}\\
\hline
FAS-TD-SF~\cite{FASTD2018arxiv} &SiW &CASIA-MFSD &39.4\\
\hline
FAS-TD-SF~\cite{FASTD2018arxiv} &CASIA-SURF &CASIA-MFSD &37.3\\
\hline
\end{tabular}
}
\label{tab:CASIA-MFSD}
\end{table}

\section{Discussion}

{\noindent \textbf{Why not collect video replay attacks?}} 
In the design stage of the proposed dataset, we found that replay videos are presented black in depth images, \ie, pixels in depth images are zero because of the same depth value for replay videos. It means that replay video attacks are easy to be recognized by means of depth data. This is why the developed dataset contains only the print attack and not the video replay attack. Besides, there are many other ways of attacking and we plan to continuously include more presentation attack ways (\eg, 3D masks).

{\noindent \textbf{Why use the ROC curve as the evaluation metric?}} 
As shown in Table~\ref{tab:ablation}, accurate results are achieved on the CASIA-SURF dataset for traditional metrics, \eg, APCER=$1.6\%$, NPCER=$0.08\%$, ACER=$0.8\%$. However, APCER=$1.6\%$ means about $2$ fake samples from $100$ attackers will be treated as real ones. This is below the accuracy requirements of real applications, \eg, face payment and phone unlock. To decrease the gap between technology development and practical applications, the ROC curve is more suitable as the evaluation metric for face anti-spoofing to reflects whether algorithms meet the requirements of a given real application.

\section{Conclusion}
This paper builds a large-scale multi-modal face anti-spoofing dataset namely CASIA-SURF. It is the largest one in terms of number of subjects, data samples, and number of visual data modalities. Comprehensive evaluation metrics, diverse evaluation protocols, training/validation/testing subsets and a measurement tool are also provided to develop a new benchmark. We believe this dataset will push the state-of-the-art in face anti-spoofing. Furthermore, we proposed a multi-modal multi-scale fusion method, which performs modality-dependent feature re-weighting to select the more informative channel features while suppressing the less informative ones for each modality across different scales. Extensive experiments have been conducted on the CASIA-SURF dataset to verify the generalization capability of models trained on the proposed dataset and the benefit of using multiple visual modalities. In the further, we plan to continuously increase the diversity of the dataset by including more presentation attack modalities (\eg, 3D masks) and more subjects (\eg, different ethnicity). On the other hand, we also plan to study heterogeneous face anti-spoofing using the cross-modal evaluation protocol.

\section{Acknowledgements}
This work has been partially supported by the Chinese National Natural Science Foundation Projects $\#$61961160704, $\#$61876179, $\#$61872367, by the Science and Technology Development Fund of Macau (Grant No.~0008/2018/A1, 0025/2019/A1, 0019/2018/ASC, 0010/2019/AFJ, 0025/2019/AKP), by the Spanish project TIN2016-74946-P (MINECO/FEDER, UE) and CERCA Programme / Generalitat de Catalunya, and by ICREA under the ICREA Academia programme. We gratefully acknowledge Surfing Technology Beijing co., Ltd (www.surfing.ai) to capture and provide us this high quality dataset for this research. 

\ifCLASSOPTIONcaptionsoff
  \newpage
\fi

\bibliographystyle{IEEEtran}
\bibliography{IEEEabrv,reference}

\begin{thebibliography}{10}
\providecommand{\url}[1]{#1}
\csname url@samestyle\endcsname
\providecommand{\newblock}{\relax}
\providecommand{\bibinfo}[2]{#2}
\providecommand{\BIBentrySTDinterwordspacing}{\spaceskip=0pt\relax}
\providecommand{\BIBentryALTinterwordstretchfactor}{4}
\providecommand{\BIBentryALTinterwordspacing}{\spaceskip=\fontdimen2\font plus
\BIBentryALTinterwordstretchfactor\fontdimen3\font minus
  \fontdimen4\font\relax}
\providecommand{\BIBforeignlanguage}[2]{{%
\expandafter\ifx\csname l@#1\endcsname\relax
\typeout{** WARNING: IEEEtran.bst: No hyphenation pattern has been}%
\typeout{** loaded for the language `#1'. Using the pattern for}%
\typeout{** the default language instead.}%
\else
\language=\csname l@#1\endcsname
\fi
#2}}
\providecommand{\BIBdecl}{\relax}
\BIBdecl

\bibitem{zhang2019refineface}
S.~Zhang, C.~Chi, Z.~Lei, and S.~Z. Li, ``Refineface: Refinement neural network
  for high performance face detection,'' \emph{arXiv}, 2019.

\bibitem{wang2019mis}
X.~Wang, S.~Zhang, S.~Wang, T.~Fu, H.~Shi, and T.~Mei, ``Mis-classified vector
  guided softmax loss for face recognition,'' in \emph{AAAI}, 2020.

\bibitem{zhang2019single}
S.~Zhang, L.~Wen, H.~Shi, Z.~Lei, S.~Lyu, and S.~Z. Li, ``Single-shot
  scale-aware network for real-time face detection,'' \emph{IJCV}, 2019.

\bibitem{chi2018selective}
C.~Chi, S.~Zhang, J.~Xing, Z.~Lei, S.~Z. Li, and X.~Zou, ``Selective refinement
  network for high performance face detection,'' in \emph{AAAI}, 2019.

\bibitem{zhang2019faceboxes}
S.~Zhang, X.~Wang, Z.~Lei, and S.~Z. Li, ``Faceboxes: A cpu real-time and
  accurate unconstrained face detector,'' \emph{Neurocomputing}, 2019.

\bibitem{wang2018ensemble}
X.~Wang, S.~Zhang, Z.~Lei, S.~Liu, X.~Guo, and S.~Z. Li, ``Ensemble soft-margin
  softmax loss for image classification,'' in \emph{IJCAI}, 2018.

\bibitem{zhang2019improved}
S.~Zhang, R.~Zhu, X.~Wang, T.~Fu, S.~Wang, T.~Mei, and S.~Z. Li, ``Improved
  selective refinement network for face detection,'' \emph{arXiv}, 2019.

\bibitem{zhang2018detecting}
S.~Zhang, X.~Zhu, Z.~Lei, X.~Wang, H.~Shi, and S.~Z. Li, ``Detecting face with
  densely connected face proposal network,'' \emph{Neurocomputing}, vol. 284,
  pp. 119--127, 2018.

\bibitem{Boulkenafet2016Face}
Z.~Boulkenafet, J.~Komulainen, and A.~Hadid, ``Face spoofing detection using
  colour texture analysis,'' \emph{TIFS}, 2016.

\bibitem{Boulkenafet2017Face}
------, ``Face antispoofing using speeded-up robust features and fisher vector
  encoding,'' \emph{SPL}, 2017.

\bibitem{Chingovska_BIOSIG-2012}
I.~Chingovska, A.~Anjos, and S.~Marcel, ``On the effectiveness of local binary
  patterns in face anti-spoofing,'' in \emph{BIOSIG}, 2012.

\bibitem{Zhang2012A}
Z.~Zhang, J.~Yan, S.~Liu, Z.~Lei, D.~Yi, and S.~Z. Li, ``A face antispoofing
  database with diverse attacks,'' in \emph{ICB}, 2012.

\bibitem{ERDOGMUS_BTAS-2013}
N.~Erdogmus and S.~Marcel, ``Spoofing in 2d face recognition with 3d masks and
  anti-spoofing with kinect,'' in \emph{BTAS}, 2014.

\bibitem{Dhamecha2014Recognizing}
T.~I. Dhamecha, R.~Singh, M.~Vatsa, and A.~Kumar, ``Recognizing disguised
  faces: Human and machine evaluation,'' \emph{Plos One}, vol.~9, 2014.

\bibitem{RaghavendraPresentation}
R.~Raghavendra, K.~B. Raja, and C.~Busch, ``Presentation attack detection for
  face recognition using light field camera,'' \emph{TIP}, 2015.

\bibitem{Wen2015Face}
D.~Wen, H.~Han, and A.~K. Jain, ``Face spoof detection with image distortion
  analysis,'' \emph{TIFS}, 2015.

\bibitem{Costa2016The}
A.~Costa-Pazo, S.~Bhattacharjee, E.~Vazquez-Fernandez, and S.~Marcel, ``The
  replay-mobile face presentation-attack database,'' in \emph{BIOSIG}, 2016.

\bibitem{liu20163d}
S.~Liu, B.~Yang, P.~C. Yuen, and G.~Zhao, ``A 3d mask face anti-spoofing
  database with real world variations,'' in \emph{CVPRW}, 2016, pp. 100--106.

\bibitem{msspoof-2015}
I.~Chingovska, N.~Erdogmus, A.~Anjos, and S.~Marcel, ``Face recognition systems
  under spoofing attacks,'' in \emph{Face Recognition Across the Imaging
  Spectrum}, 2016.

\bibitem{steiner2016reliable}
H.~Steiner, A.~Kolb, and N.~Jung, ``Reliable face anti-spoofing using
  multispectral swir imaging,'' in \emph{ICB}, 2016, pp. 1--8.

\bibitem{HolgerDesign}
S.~Holger, S.~Sebastian, K.~Andreas, and J.~Norbert, ``Design of an active
  multispectral swir camera system for skin detection and face verification,''
  \emph{Journal of Sensors}, vol. 2016, pp. 1--16, 2016.

\bibitem{raghavendra2017vulnerability}
R.~Raghavendra, K.~B. Raja, S.~Venkatesh, F.~A. Cheikh, and C.~Busch, ``On the
  vulnerability of extended multispectral face recognition systems towards
  presentation attacks,'' in \emph{ISBA}, 2017, pp. 1--8.

\bibitem{manjani2017detecting}
I.~Manjani, S.~Tariyal, M.~Vatsa, R.~Singh, and A.~Majumdar, ``Detecting
  silicone mask-based presentation attack via deep dictionary learning,''
  \emph{TIFS}, vol.~12, no.~7, pp. 1713--1723, 2017.

\bibitem{agarwal2017face}
A.~Agarwal, D.~Yadav, N.~Kohli, R.~Singh, M.~Vatsa, and A.~Noore, ``Face
  presentation attack with latex masks in multispectral videos,'' in
  \emph{CVPRW}, 2017, pp. 81--89.

\bibitem{Boulkenafet2017OULU}
Z.~Boulkenafet, J.~Komulainen, L.~Li, X.~Feng, and A.~Hadid, ``Oulu-npu: A
  mobile face presentation attack database with real-world variations,'' in
  \emph{FG}, 2017.

\bibitem{Liu2018Learning}
Y.~Liu, A.~Jourabloo, and X.~Liu, ``Learning deep models for face
  anti-spoofing: Binary or auxiliary supervision,'' in \emph{CVPR}, 2018.

\bibitem{george2019biometric}
A.~George, Z.~Mostaani, D.~Geissenbuhler, O.~Nikisins, A.~Anjos, and S.~Marcel,
  ``Biometric face presentation attack detection with multi-channel
  convolutional neural network,'' \emph{TIFS}, 2019.

\bibitem{Jourabloo2018Face}
A.~Jourabloo, Y.~Liu, and X.~Liu, ``Face de-spoofing: Anti-spoofing via noise
  modeling,'' \emph{arXiv}, 2018.

\bibitem{deng2009imagenet}
J.~Deng, W.~Dong, R.~Socher, L.-J. Li, K.~Li, and L.~Fei-Fei, ``Imagenet: A
  large-scale hierarchical image database,'' in \emph{CVPR}, 2009.

\bibitem{yi2014learning}
D.~Yi, Z.~Lei, S.~Liao, and S.~Z. Li, ``Learning face representation from
  scratch,'' \emph{arXiv}, 2014.

\bibitem{liu2017sphereface}
W.~Liu, Y.~Wen, Z.~Yu, M.~Li, B.~Raj, and L.~Song, ``Sphereface: Deep
  hypersphere embedding for face recognition,'' in \emph{CVPR}, 2017.

\bibitem{wang2018support}
X.~Wang, S.~Wang, S.~Zhang, T.~Fu, H.~Shi, and T.~Mei, ``Support vector guided
  softmax loss for face recognition,'' \emph{arXiv}, 2018.

\bibitem{DBLP:conf/cvpr/abs-1812-00408}
S.~Zhang, X.~Wang, A.~Liu, C.~Zhao, J.~Wan, S.~Escalera, H.~Shi, Z.~Wang, and
  S.~Z. Li, ``A dataset and benchmark for large-scale multi-modal face
  anti-spoofing,'' in \emph{CVPR}, 2019.

\bibitem{Bhattacharjee_BTAS2018_2018}
S.~Bhattacharjee, A.~Mohammadi, and S.~Marcel, ``Spoofing deep face recognition
  with custom silicone masks,'' in \emph{BTAS}, 2018.

\bibitem{Kim2009Masked}
Y.~Kim, J.~Na, S.~Yoon, and J.~Yi, ``Masked fake face detection using radiance
  measurements,'' \emph{JOSA A}, 2009.

\bibitem{Kose2013Countermeasure}
N.~Kose and J.-L. Dugelay, ``Countermeasure for the protection of face
  recognition systems against mask attacks,'' in \emph{FG}, 2013.

\bibitem{Pan2007Eyeblink}
G.~Pan, L.~Sun, Z.~Wu, and S.~Lao, ``Eyeblink-based anti-spoofing in face
  recognition from a generic webcamera,'' in \emph{ICCV}, 2007.

\bibitem{wang2009face}
L.~Wang, X.~Ding, and C.~Fang, ``Face live detection method based on
  physiological motion analysis,'' \emph{TST}, 2009.

\bibitem{kollreider2008verifying}
K.~Kollreider, H.~Fronthaler, and J.~Bigun, ``Verifying liveness by multiple
  experts in face biometrics,'' in \emph{CVPR Workshops}, 2008.

\bibitem{Bharadwaj2013Computationally}
S.~Bharadwaj, T.~I. Dhamecha, M.~Vatsa, and R.~Singh, ``Computationally
  efficient face spoofing detection with motion magnification,'' in
  \emph{CVPR}, 2013.

\bibitem{Pan2011Monocular}
G.~Pan, L.~Sun, Z.~Wu, and Y.~Wang, ``Monocular camera-based face liveness
  detection by combining eyeblink and scene context,'' \emph{TCS}, 2011.

\bibitem{Komulainen2014Context}
J.~Komulainen, A.~Hadid, and M.~Pietikainen, ``Context based face
  anti-spoofing,'' in \emph{BTAS}, 2013.

\bibitem{Wang2013Face}
T.~Wang, J.~Yang, Z.~Lei, S.~Liao, and S.~Z. Li, ``Face liveness detection
  using 3d structure recovered from a single camera,'' in \emph{ICB}, 2013.

\bibitem{De2012Moving}
M.~De~Marsico, M.~Nappi, D.~Riccio, and J.-L. Dugelay, ``Moving face spoofing
  detection via 3d projective invariants,'' in \emph{ICB}, 2012.

\bibitem{Kim2013Face}
S.~Kim, S.~Yu, K.~Kim, Y.~Ban, and S.~Lee, ``Face liveness detection using
  variable focusing,'' in \emph{ICB}, 2013.

\bibitem{Li2004Live}
J.~Li, Y.~Wang, T.~Tan, and A.~K. Jain, ``Live face detection based on the
  analysis of fourier spectra,'' \emph{BTHI}, 2004.

\bibitem{chingovska2012effectiveness}
I.~Chingovska, A.~Anjos, and S.~Marcel, ``On the effectiveness of local binary
  patterns in face anti-spoofing,'' in \emph{BIOSIG}, 2012.

\bibitem{Yang2013Face}
J.~Yang, Z.~Lei, S.~Liao, and S.~Z. Li, ``Face liveness detection with
  component dependent descriptor.'' in \emph{ICB}, 2013.

\bibitem{Maatta2012Face}
J.~Maatta, A.~Hadid, and M.~Pietikainen, ``Face spoofing detection from single
  images using texture and local shape analysis,'' \emph{IET biometrics}, 2012.

\bibitem{schwartz2011face}
W.~R. Schwartz, A.~Rocha, and H.~Pedrini, ``Face spoofing detection through
  partial least squares and low-level descriptors,'' in \emph{IJCB}, 2011.

\bibitem{tronci2011fusion}
R.~Tronci, D.~Muntoni, G.~Fadda, M.~Pili, N.~Sirena, G.~Murgia, M.~Ristori,
  S.~Ricerche, and F.~Roli, ``Fusion of multiple clues for photo-attack
  detection in face recognition systems,'' in \emph{IJCB}, 2011.

\bibitem{Pereira2013Can}
T.~de~Freitas~Pereira, A.~Anjos, J.~M. De~Martino, and S.~Marcel, ``Can face
  anti-spoofing countermeasures work in a real world scenario?'' in \emph{ICB},
  2013.

\bibitem{komulainen2013complementary}
J.~Komulainen, A.~Hadid, M.~Pietik{\"a}inen, A.~Anjos, and S.~Marcel,
  ``Complementary countermeasures for detecting scenic face spoofing attacks,''
  in \emph{ICB}, 2013.

\bibitem{feng2016integration}
L.~Feng, L.-M. Po, Y.~Li, X.~Xu, F.~Yuan, T.~C.-H. Cheung, and K.-W. Cheung,
  ``Integration of image quality and motion cues for face anti-spoofing: A
  neural network approach,'' \emph{JVCIR}, 2016.

\bibitem{li2016original}
L.~Li, X.~Feng, Z.~Boulkenafet, Z.~Xia, M.~Li, and A.~Hadid, ``An original face
  anti-spoofing approach using partial convolutional neural network,'' in
  \emph{IPTA}, 2016.

\bibitem{Patel2016Secure}
K.~Patel, H.~Han, and A.~K. Jain, ``Secure face unlock: Spoof detection on
  smartphones,'' \emph{TIFS}, 2016.

\bibitem{yang2014learn}
J.~Yang, Z.~Lei, and S.~Z. Li, ``Learn convolutional neural network for face
  anti-spoofing,'' \emph{arXiv}, 2014.

\bibitem{dlib09}
D.~E. King, ``Dlib-ml: A machine learning toolkit,'' \emph{JMLR}, 2009.

\bibitem{DBLP:conf/eccv/FengWSWZ18}
Y.~Feng, F.~Wu, X.~Shao, Y.~Wang, and X.~Zhou, ``Joint 3d face reconstruction
  and dense alignment with position map regression network,'' in \emph{ECCV},
  2018.

\bibitem{he2016deep}
K.~He, X.~Zhang, S.~Ren, and J.~Sun, ``Deep residual learning for image
  recognition,'' in \emph{CVPR}, 2016.

\bibitem{hu2018senet}
J.~Hu, L.~Shen, and G.~Sun, ``Squeeze-and-excitation networks,'' in
  \emph{CVPR}, 2018.

\bibitem{DBLP:conf/iccv/SunSSG17}
C.~Sun, A.~Shrivastava, S.~Singh, and A.~Gupta, ``Revisiting unreasonable
  effectiveness of data in deep learning era,'' in \emph{ICCV}, 2017.

\bibitem{FASTD2018arxiv}
Z.~Wang, C.~Zhao, Y.~Qin, Q.~Zhou, and Z.~Lei, ``Exploiting temporal and depth
  information for multi-frame face anti-spoofing,'' \emph{arXiv}, 2018.

\bibitem{Boulkenafet2017A}
Z.~Boulkenafet, J.~Komulainen, Z.~Akhtar, A.~Benlamoudi, and A.~Hadid, ``A
  competition on generalized software-based face presentation attack detection
  in mobile scenarios,'' in \emph{IJCB}, 2017.

\bibitem{yang2019face}
X.~Yang, W.~Luo, L.~Bao, Y.~Gao, D.~Gong, S.~Zheng, Z.~Li, and W.~Liu, ``Face
  anti-spoofing: Model matters, so does data,'' in \emph{CVPR}, 2019.

\bibitem{pinto2015face}
A.~Pinto, H.~Pedrini, W.~R. Schwartz, and A.~Rocha, ``Face spoofing detection
  through visual codebooks of spectral temporal cubes,'' \emph{TIP}, 2015.

\end{thebibliography}

\begin{IEEEbiography}[{\includegraphics[width=1in,height=1.25in,clip,keepaspectratio]{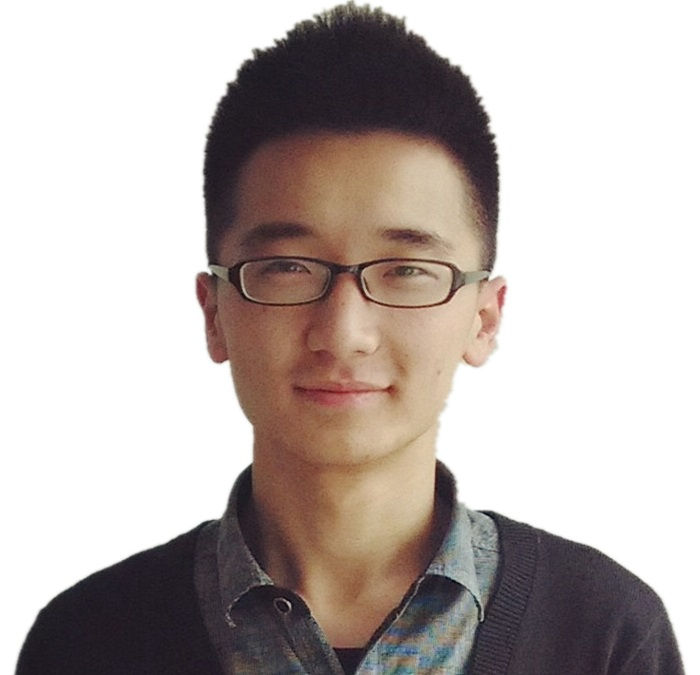}}]{Shifeng Zhang}
received the B.S. degree from the University of Electronic Science and Technology of China (UESTC), in 2015. Since September 2015, he has been a Ph.D. candidate at the National Laboratory of Pattern Recognition (NLPR), Institute of Automation, Chinese Academy of Science (CASIA). His research interests include computer vision, pattern recognition, especially with a focus on object detection, face detection, pedestrian detection, video detection.
\end{IEEEbiography}

\begin{IEEEbiography}[{\includegraphics[width=1in,height=1.25in,clip,keepaspectratio]{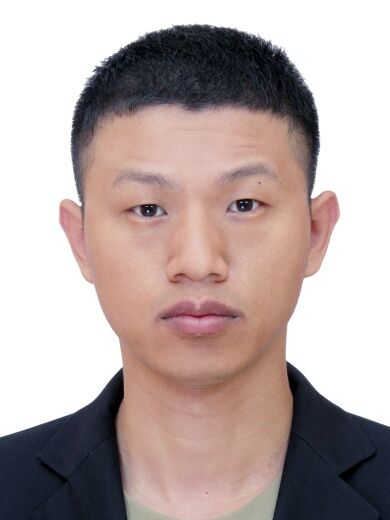}}]{Ajian Liu}
Ajian Liu received the B.E. degree from the College of Physics and Information Engineering, Shanxi Normal University, Shanxi, China in 2015, and the Master degree from the College of Information and Computer, Taiyuan University of Technology, Shanxi, China, in 2018. He is currently working toward for the PhD degree at the Faculty of Information Technology, Macau University of Science and Technology (M.U.S.T.). His main research interests include deep learning and face anti-spoofing.
\end{IEEEbiography}

\begin{IEEEbiography}[{\includegraphics[width=1in,height=1.25in,clip,keepaspectratio]{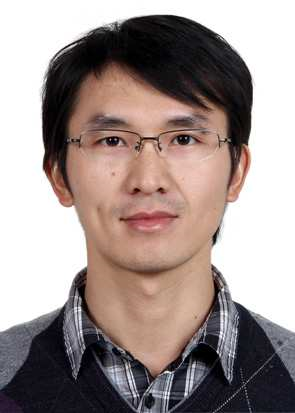}}]{Jun Wan}
received the BS degree from the China University of Geosciences, Beijing, China, in 2008, and the PhD degree from the Institute of Information Science, Beijing Jiaotong University, Beijing, China, in 2015. Since January 2015, he has been a Faculty Member with the National Laboratory of Pattern Recognition (NLPR), Institute of Automation, Chinese Academy of Science (CASIA), China, where he currently serves as an Associate Professor. His main research interests include computer vision, machine learning, especially for gesture and action recognition, facial attribution analysis.
\end{IEEEbiography}

\begin{IEEEbiography}[{\includegraphics[width=1in,height=1.25in,clip,keepaspectratio]{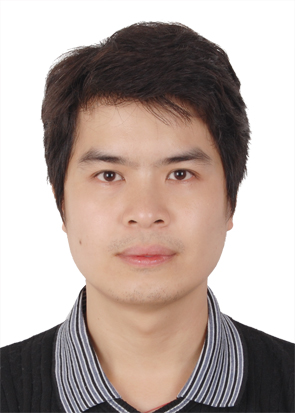}}]{Yanyan Liang}
received the B.S. degree from the Chongqing University of Communication and Posts, Chongqing, China, in 2004 and the M.S. and
Ph.D. degrees from the Macau University of Science and Technology (MUST), Macau, China, in 2006 and 2009, respectively. He is currently an Assistant Professor with MUST. He has published over 30 papers related to pattern recognition, image processing, and computer version. He is also researching on smart city applications with computer vision. His current research interests include computer vision, image processing, and machine learning.
\end{IEEEbiography}

\begin{IEEEbiography}[{\includegraphics[width=1in,height=1.25in,clip,keepaspectratio]{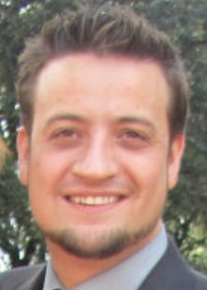}}]{Sergio Escalera}
obtained the best P.h.D. award in 2018 at Computer Vision Center, UAB. He leads the Human Pose Recovery and Behavior Analysis Group (HUPBA). He is an associate professor at the Department of Mathematics and Informatics, Universitat de Barcelona (UB). He is also a member of the Computer Vision Center (CVC) at UAB. He is vice-president of ChaLearn Challenges in Machine Learning and chair of IAPR TC-12: Multimedia and visual information systems. He has been awarded with ICREA Academia. His research interests include the visual analysis of humans, with special interest in affective and personality computing.
\end{IEEEbiography}

\begin{IEEEbiography}[{\includegraphics[width=1in,height=1.25in,clip,keepaspectratio]{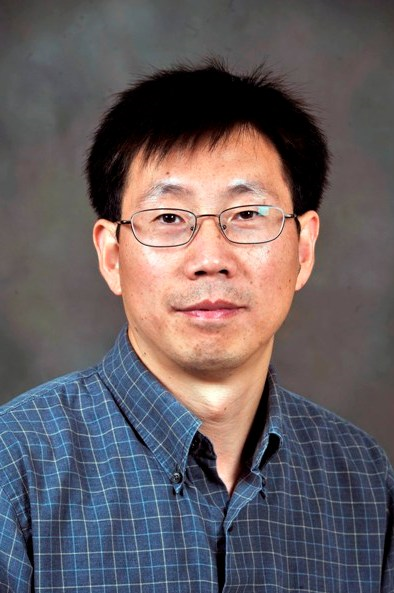}}]{Guodong Guo}
received the B.E. degree in automation from Tsinghua University, Beijing, China, the Ph.D. degree in pattern recognition and intelligent control from Chinese Academy of Sciences, Beijing, China, and the Ph.D. degree in computer science from University of Wisconsin-Madison, Madison, WI, USA. He is an Associate Professor with the Department of Computer Science and Electrical Engineering, West Virginia University (WVU), Morgantown, WV, USA. In the past, he visited and worked in several places, including INRIA, Sophia Antipolis, France; Ritsumeikan University, Kyoto, Japan; Microsoft Research, Beijing, China; and North Carolina Central University. He authored a book, {\em Face, Expression, and Iris Recognition Using Learning-based Approaches} (2008), co-edited two books, {\em Support Vector Machines Applications} (2014) and {\em Mobile Biometrics} (2017), and published over 100 technical papers. His research interests include computer vision, biometrics, machine learning, and multimedia. He received the North Carolina State Award for Excellence in Innovation in 2008, Outstanding Researcher (2017-2018, 2013-2014) at CEMR, WVU, and New Researcher of the Year (2010-2011) at CEMR, WVU. He was selected the ``People's Hero of the Week'' by BSJB under Minority Media and Telecommunications Council (MMTC) in 2013. Two of his papers were selected as ``The Best of FG'13" and ``The Best of FG'15", respectively.
\end{IEEEbiography}

\begin{IEEEbiography}[{\includegraphics[width=1in,height=1.25in,clip,keepaspectratio]{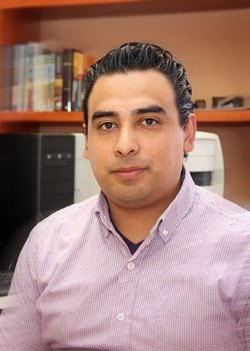}}]{Hugo Jair Escalante}
is researcher scientist at Instituto Nacional de Astrofisica, Optica y Electronica, INAOE, Mexico. 
and a director of ChaLearn, a nonprofit organization dedicated to organizing challenges, since 2011. He has been involved in the organization of several challenges in computer vision and automatic machine learning. He is reviewer at JMLR, PAMI. 
He has served as competition chair and area chair of venues like NeurIPS, PAKDD, IJCNN, among others.  
\end{IEEEbiography}

\begin{IEEEbiography}[{\includegraphics[width=1in,height=1.25in,clip,keepaspectratio]{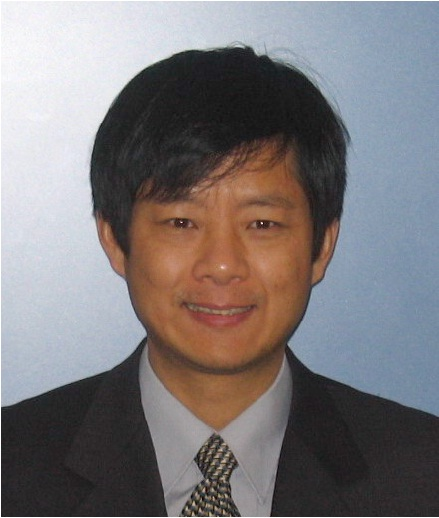}}]{Stan Z. Li}
received the B.Eng degree from Hunan University, the M.Eng degree from National University of Defense Technology, and the Ph.D. degree from Surrey University. He is currently a professor and the director of Center for Biometrics and Security Research (CBSR), Institute of Automation, Chinese Academy of Sciences (CASIA). He was with MSRA as a researcher from 2000 to 2004. Prior to that, he was an associate professor in the Nanyang Technological University. His research interests include image and vision processing, pedestrian recognition and biometrics. He has published more than 300 papers in international journals and conferences, and authored and edited 8 books. He was an associate editor of the IEEE TPAMI and is acting as the editor-in-chief for the Encyclopedia of Biometrics. He served as a program co-chair for ICB 2007, 2009, 2013, 2014, 2015, 2016 and 2018, and has been involved in organizing other international conferences and workshops in the fields of his research interest. He was elevated to IEEE fellow for his contributions to the fields of pedestrian recognition, pattern recognition and computer vision and he is a member of the IEEE Computer Society.
\end{IEEEbiography}

\end{document}